\documentclass[11pt]{article}
\pdfoutput=1

\usepackage{arXiv}

\usepackage[utf8]{inputenc} 
\usepackage[T1]{fontenc}    
\usepackage{hyperref}       
\usepackage{url}            
\usepackage{booktabs}       
\usepackage{amsfonts}       
\usepackage{nicefrac}       
\usepackage{microtype}      
\usepackage{lipsum}
\usepackage{graphicx}
\usepackage{color}
\usepackage{enumitem}
\usepackage{amsmath}
\usepackage{amssymb}
\usepackage{multicol}
\usepackage{multirow}
\usepackage{algorithm}
\usepackage{algorithmic}
\usepackage{bm}
\usepackage{array}
\usepackage{subfigure}
\usepackage{subfiles}
\usepackage{epstopdf}

\newtheorem{defn}{Definition}

\graphicspath{{images/}}

\title{A Hybrid Systems-based Hierarchical Control Architecture for Heterogeneous Field Robot Teams}

\author{
  Chanyoung Ju and Hyoung Il Son\thanks{Corresponding author.} \\
  Department of Rural and Biosystems Engineering\\
  Chonnam National University\\
  77 Yongbong-ro, Buk-Gu, Gwnagju 61186, Repulbic of Korea \\
  \texttt{cksdud15@gmail.com, hison@jnu.ac.kr} \\
}

\begin{document}
\maketitle

\begin{abstract}
Field robot systems have recently been applied to a wide range of research fields. Making such systems more automated, advanced, and activated requires cooperation among heterogeneous robots. Classic control theory is inefficient in managing large-scale complex dynamic systems. Therefore, the supervisory control theory based on discrete event system needs to be introduced to overcome this limitation. In this study, we propose a hybrid systems-based hierarchical control architecture through a supervisory control-based high-level controller and a traditional control-based low-level controller. The hybrid systems and its dynamics are modeled through a formal method called hybrid automata, and the behavior specifications expressing the control objectives for cooperation are designed. Additionally, a modular supervisor that is more scalable and maintainable than a centralized supervisory controller was synthesized. The proposed hybrid systems and hierarchical control architecture were implemented, validated, and then evaluated for performance through the physics-based simulator. Experimental results confirmed that the heterogeneous field robot team satisfied the given specifications and presented systematic results, validating the efficiency of the proposed control architecture.
\end{abstract}


\section{Introduction}
\label{sec:1}

In recent years, field robots have been widely deployed in nuclear plants, agriculture~\cite{8782102} and construction, and underwater~\cite{8809889}, as well as military areas, and their usefulness has attracted a considerable amount of attention from environmental scientists and robotics engineers. However, field robotics research is complicated because agents need to perform tasks or missions in outdoor, unstructured, and unknown environments. Nevertheless, various types of robots (e.g., unmanned ground vehicles (UGVs), unmanned aerial vehicles (UAVs), mobile manipulators, humanoids, and hybrid UAVs~\cite{ozdemir2014design}) are being developed to cope with disasters and replace labor (e.g., by automating agricultural work, such as spraying, cultivation, grafting, and harvesting)~\cite{schwarz2017nimbro,christiansen2017designing,silwal2017design}. However, issues of application and commercialization have gradually emerged owing to the considerable gap between the current robot technology and the required technology in the real field. In particular, the development of control systems for the cooperation of homogeneous and heterogeneous field robots is still an enormous challenge.

For the eventual autonomy of temporal–spatial missions (e.g., mapping, sensing, sampling, and monitoring) using field robots, novel control frameworks that enable multiple heterogeneous robots to collaborate must be introduced. However, the classical control theory based on differential equations has some limitations in handling large-scale complex dynamic systems (e.g., heterogeneous field robots). For example, multiple UAV systems for remote sensing~\cite{ju2018multiple}, autonomous underwater vehicles (AUVs) for distributed formation tracking~\cite{8051276} and path planning~\cite{7345594}, and heterogeneous multi-robot systems for exploration, sampling~\cite{manjanna2018heterogeneous}, and mapping~\cite{roldan2016heterogeneous} have been studied through the traditional control approach. These approaches are inadequate for handling reactivity, scalability, maintainability, modularity, and systematic analysis when field robots are added to or removed from a team, or when a given mission is modified (e.g., a change of tasks in local or global areas), even if robustness and adaptiveness are considered in the controller. Additionally, the field robot systems need to meet the practicality requirements of a dynamic environment, which reflects asynchronous events.

One approach to solve this problem involves the use of discrete event systems (DESs) and formal methods to systematically analyze the states, events, and behaviors of large-scale dynamic systems and design powerful controllers~\cite{cassandras2009introduction}. The formal methods typically include automata, Petri nets, temporal logic, behavior trees, and min-max algebra. Recently, supervisory control theory (SCT), also known as the Ramadge-Wonham framework (RW framework), based on automata has become an intensive approach to controlling DES behavior~\cite{wonham2015supervisory}. SCT has proven to be efficient for large-scale dynamic systems as this approach can design supervisors that meet behavior specifications of plants and maximally enables eligible events~\cite{son2011design}. Consequently, SCT-based control systems have been developed for sumo robots~\cite{torrico2016modeling}, industrial UGVs~\cite{gonzalez2017supervisory}, multicopters~\cite{quan2017failsafe}, patient support tables~\cite{theunissen2013application}, and fuzzy DESs~\cite{jayasiri2011behavior}. However, the challenging problems that arise in this domain are how to debug, use, and apply formal methods for dynamic systems. More importantly, automata-based SCT, which is a rule-based approach, requires an implementable solution without event generators because the supervisory controller enables and disables controllable events while observing events occurring in the target plant.

Therefore, a new approach using DES and SCT is needed for the explicit cooperation of heterogeneous field robots. In this study, we propose a hybrid systems-based hierarchical control (HSHC) architecture that combines classical control theory based on a continuous time system (CTS) and DES-based SCT to overcome these weaknesses. Because HSHC tries to integrate the advantages of each system and control theory, it also considerably overcomes the existing limitations (i.e., computational complexity, modeling uncertainties, or differences of dynamics) of large-scale complex dynamic systems with unstructured and unknown environments. Additionally, it has the advantage of systematically analyzing the system behaviors (i.e., state transitions), events occurrence and sequence, and control behaviors in any condition. Hence, HSHC is an ideal control architecture for heterogeneous field robot teams where system modeling and controller design are extremely challenging.

\begin{figure}[tb]
\centering
\includegraphics[trim={2.0cm 0 2.0cm 0}, width=0.6\linewidth]{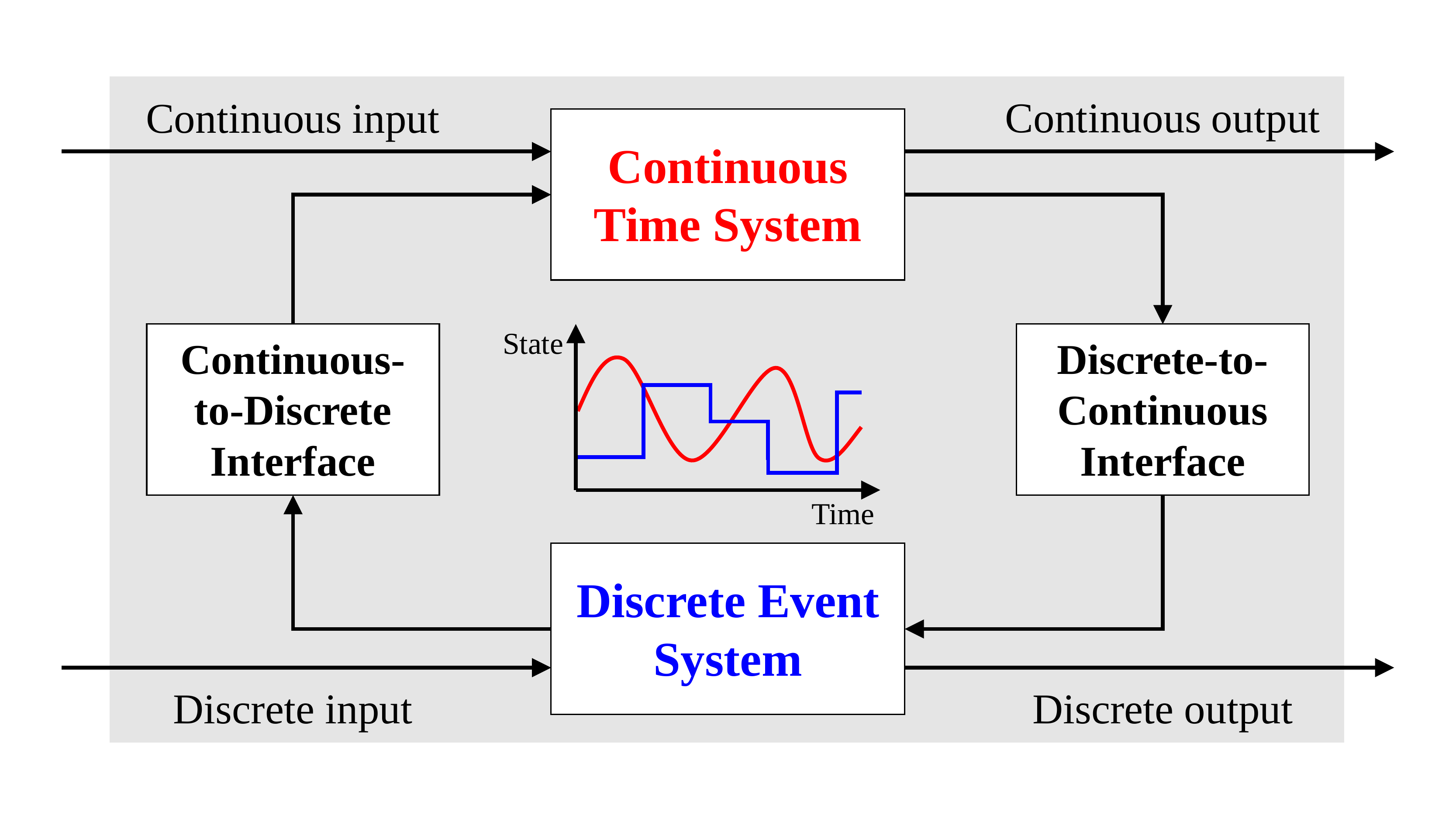}
\caption{Concept of hybrid systems consisting of CTS and DES.}
\label{fig:hybrid_system}
\end{figure}

\begin{figure}[tb]
\centering
\includegraphics[trim={2.0cm 0 2.0cm 0}, width=0.6\linewidth]{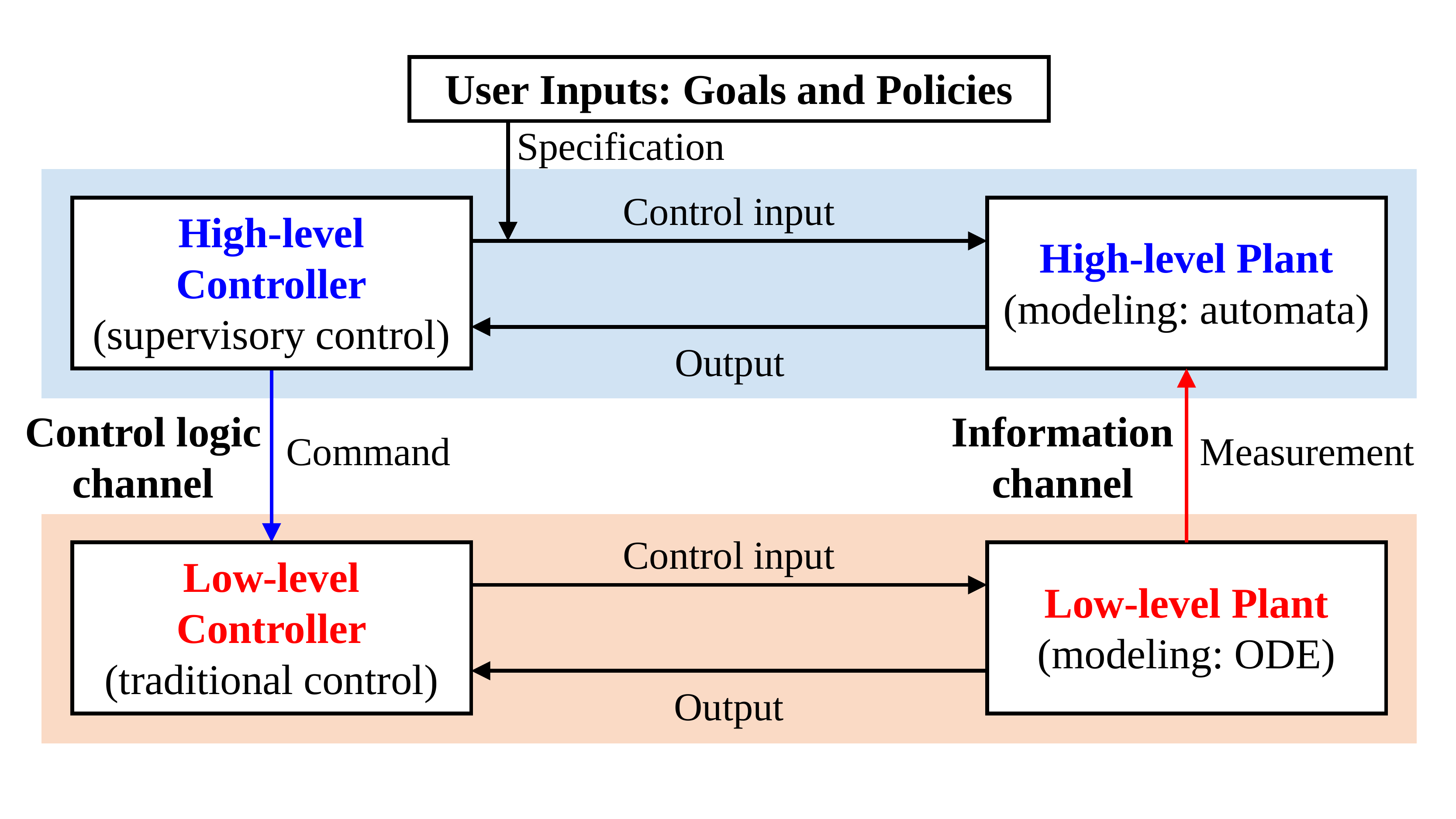}
\caption{Concept of hierarchical control architecture composed of a DES-based high-level controller and a CTS-based low-level controller. (ODE stands for ordinary differential equation.)}
\label{fig:hierarchical_control}
\end{figure}

\subsection{Objective of the Study}
This study aims to propose an HSHC architecture for heterogeneous field robot cooperation. The hybrid systems are the coupling scheme consisting of a CTS and a DES, as shown in Fig.~\ref{fig:hybrid_system}. The hierarchical control architecture is illustrated in Fig.~\ref{fig:hierarchical_control}. We design the specification language representing the goals and policies, high-level plant, and high-level controller based on the DES. However, when controlling the robot system with dynamics taken into consideration, a low-level controller is imperatively needed. Therefore, we model the plant based on the CTS and design the low-level controller using ordinary differential equations. The command of a supervisor is transmitted to the low-level controller through the control logic channel, and the output of the field robot plant (measured values, state variables, occurred events, etc.) is transmitted to the high-level stage through the information channel. Consequently, the supervisory controller can feedback control the hybrid systems by selectively enabling controllable events based on this information.

In our previous work~\cite{8834867,ju2019hybrid}, we also designed a centralized supervisory controller to control a heterogeneous agricultural field robot system. Additionally, the finite-state automata theory was applied to model the entire system, focusing only on the behavior of field robots for collaboration. In other words, it was a preliminary study to verify the feasibility of HSHC that we propose here. In this study, we modeled the field robot system based on the dynamics of each mobile robot, and extended the previous research using the hybrid automata and hybrid systems. Furthermore, the modular supervisor was designed to be more scalable, maintainable, and superior to the centralized supervisory controller to manage large-scale dynamic systems. In summary, the goals of our research are to develop the HSHC system that can systematically model, control, and analyze heterogeneous field robots based on a novel approach. The proposed hybrid systems and modular supervisors are implemented, experimented upon, validated, and evaluated in a physics-based simulator similar to a field environment, and we present its systematic results and effectiveness.

\subsection{Related Works}
In this section, we review related studies based on formal methods for multi-robots, in particular, DES and SCT. SCT, a theory proposed by Ramadge and Wonham in 1987, introduced a methodology for controlling DESs~\cite{ramadge1987supervisory}. It has been applied mainly to the field of manufacturing automation~\cite{son2007failure}, but recently it has been introduced into a cyber-physical robotic system (i.e., a dynamic system). For task allocation for cooperative robot teams, the supervisory controller was designed to accommodate flexibility in task assignments, robot coordination, and tolerance of failures and repairs~\cite{tsalatsanis2012dynamic}. By designing a control architecture for the application of heterogeneous multi-robot teams to urban search and rescue and presenting its simulation results, Yugang \textit{et al}. \cite{liu2015supervisory} found that this approach was effective for multi-robot control in field applications and is robust to varying scene scenarios and increasing team size. Nevertheless, most research on the supervisory control of multiple robots aims at autonomous navigation~\cite{dulce2019autonomous} through behavior coordination~\cite{mendiburu2016behavior} and formation control~\cite{5354788}. However, these works focus on how to model the DES and how to analyze the supervisory controller rather than the implementation or demonstration of their robot system. Examples of implementation and application of SCT include warehouse automation~\cite{tatsumoto2018application}, agricultural UAVs~\cite{ju2018discrete}, exploration~\cite{dai2016cooperative}, theme parking vehicles~\cite{forschelen2012application}, swarm robotics~\cite{lopes2016supervisory}, and attempts to integrate self-development tools and frameworks~\cite{jordan2017educational,pinheiro2015nadzoru,rahmani2018spectr,hill2017scaling,malik2017supremica}. Most of the aforementioned studies employ a simple supervisory control through an automata model and do not handle the implementation problem by combining with the CTS. In recent research, for example, a learning-based synthesis approach~\cite{8259248}, a probabilistic DES under partial observation~\cite{8667731}, fuzzy DES-based shared control~\cite{7247698}, robust supervisory-based control~\cite{10.1007/978-3-319-08338-4_10}. and a multilevel DES for bus structure~\cite{8795835} have been studied for application to the robotics field based on DES and SCT, but these are still in a simple simulation stage. These studies, in which supervisory control and a dynamic system based on DES are proposed, do not guarantee solution to the above-mentioned problems. Therefore, additional research is required to control, implement, and sustain large-scale dynamic systems by using DES and SCT beyond the traditional control method in field robotics. To the best of our knowledge, additionally, no prior studies have examined the supervisory control of heterogeneous field robots. We address this issue by proposing an HSHC architecture through a new perspective on supervisory control and classic control.

\subsection{Structure of the Paper}
The remainder of this paper is organized as follows: In Section \ref{sec:2}, we present the process of designing a CTS-based low-level controller and distributed swarm control algorithm, consisting of obstacle avoidance control, formation control, and path-following control.
In Section \ref{sec:3}, we introduce the overall DES concept and a system modeling method using automata and describe the design of a high-level controller through SCT and detail the conditions for a proper supervisor. In Section \ref{sec:4}, we describe the control approach proposed in this study, HSHC architecture, method of combining CTS and DES, hybrid automata-based system modeling, design of behavior specifications to achieve control objectives, and synthesis of the modular supervisor for heterogeneous field robots. In Section \ref{sec:5}, we present the implementation procedure, experimental setup, and experimental results through a physics-based simulator to evaluate the proposed control system. Finally, in Section \ref{sec:6}, we summarize the conclusion and address the direction of future research.

\section{CTS-based Low-level Control}
\label{sec:2}

\subsection{Ordinary Differential Equation-based System Modeling}
\label{sec:2.1}

We use the dynamic UAV and UGV models to design a low-level controller based on a CTS. The kinematic and dynamic equations for each vehicle are described in Sections \ref{sec:2.1.1} and \ref{sec:2.1.2}.

\subsubsection{UAV model}
\label{sec:2.1.1}

We consider $N$ UAVs with three degrees of freedom (DOF). The positions are marked by $p_i \in \Re^3$, $i=1,2,\dots,N$. The flight control input of UAVs is derived from the following dynamics and kinematic equations:
\begin{equation}
    m_i\ddot{p}_i = -\lambda_i\mathcal{R}_{i}e_3+m_ige_3 + \delta_i \label{eq:mi}
\end{equation}
\begin{equation}
    \mathcal{J}_{i}\dot{w}_i + S(w_i)\mathcal{J}_{i}w_i = \gamma_i + \eta_i \label{eq:ji}
\end{equation}
\begin{equation}
    \dot{R}_i = \mathcal{R}_{i}S(w_i)   \label{eq:ri}
\end{equation}
where $m_i>0$ denotes the mass, $p_i:=[p_1;p_2;\ldots;p_N]\in\Re^{3N}$ represents the Cartesian center-of-mass position represented in the northeast (NE) inertial frame $\{O\}$, $\lambda_i \in \Re$ indicates the thrust control input along $e_3$ (with $e_1$, $e_2$, and $e_3$ indicating N, E, and D directions), $\mathcal{R}_{i} \in SO(3)$ presents the rotational matrix describing the body frame $\{B\}$ of the UAV w.r.t. the inertial frame $\{O\}$, $\mathcal{A}$ is the gravitational constant, $e_3=[0,0,1]^T $ denotes the basis vector representing the down direction and representing that thrust and gravity act in the D direction, $\mathcal{J}_{i} \in \Re^{3 \times 3}$ indicates the inertia matrix w.r.t. the body frame $\{B\}$, $w_i \in \Re^3$ presents the angular speed of the UAV relative to the inertial frame $\{O\}$ represented in the body frame $\{B\}$, $\gamma_i \in \Re^3$ represents the attitude torque control input, $\delta_i, \eta_i \in \Re^3$ denote the aerodynamic perturbations, and $\Upsilon(\diamond):\Re^3 \rightarrow so(3)$ denotes the skew-symmetric operator defined subject to $\alpha, \beta \in \Re^3,\; \Upsilon(\alpha)\beta = \alpha \times \beta$. For typical UAV flying, $\delta_i,\eta_i \approx 0$.

The relationships among angular speed, thrust, and torque of each propeller for low-level control input are as follows:
\begin{equation}
    \begin{pmatrix}
    \lambda_i \\ \gamma_{i1} \\ \gamma_{i2} \\ \gamma_{i3} 
    \end{pmatrix}
    =
    \begin{bmatrix}
    \kappa & \kappa & \kappa & \kappa \\ 0 & -L & 0 & L \\ L & 0 & -L & 0 \\ \beta & -\beta & \beta & -\beta
    \end{bmatrix}
    \begin{pmatrix}
    \omega_{i1}^2 \\ \omega_{i2}^2 \\ \omega_{i3}^2 \\ \omega_{i4}^2 
    \end{pmatrix}
    \label{eq:aaa}
\end{equation}
where $\kappa$ and $\beta$ are constants that determine the relationship among angular speed, thrust, and torque; $L$ is the distance between the propeller and the center of the UAV; $\gamma_{ij}$ denotes the torque in the roll, pitch, and yaw directions; and $\omega_{ij}$ denotes the angular speed of each propeller. As a result, when the desired thrust and torque are determined, the desired control input can be generated by controlling the motor through equation (\ref{eq:aaa}).

\subsubsection{UGV model}
\label{sec:2.1.2}

We consider $N$ UGVs with the positions when the coordinate frame is $\{U\}$ and the body frame is $\{D\}$. The kinematic and dynamics equations of the UGV are as follows:
\begin{equation} \label{eq:ugv_kine}
\begin{split}
    \dot{x_i} &= \nu_i\cos(\psi_i)\\
    \dot{y_i} &= \nu_i\sin(\psi_i)\\
    \dot{\psi_i} &= \omega_i
\end{split}
\end{equation}

\begin{equation}
    \begin{split}
        D(x_i) \begin{pmatrix} \dot{\nu_i} \\ \dot{\omega_i} \end{pmatrix} + Q(x_i,w_i) \begin{pmatrix} \nu_i \\ \dot{\psi_i} \end{pmatrix} = \mu_i + \varrho_i \label{eq:ugv}
    \end{split}
\end{equation}
where $x_i$, $y_i$ are the coordinates of the UGV w.r.t. coordinate frame $\{U\}$; $\psi_i$ denotes the heading angle; $\nu_i$ and $\omega_i$ are the linear and angular speeds of the $i$th robot expressed in $\{D\}$, respectively; $\mu_i=[\mu_i^\nu,\mu_i^\omega]^T$ and $\varrho_i=[\varrho_i^\nu,\varrho_i^\omega]^T$ are the control input and the external force and torque, respectively; $D(x_i)\in\Re^3$ is the positive-definite symmetric inertia matrix; and $Q(x_i,\omega_i) \in \Re^{3 \times 3}$ is the Coriolis matrix.

\begin{figure}[tb]
\centering
\includegraphics[trim={1.0cm 0.0cm 1.0cm 0.0cm}, width=0.6\linewidth]{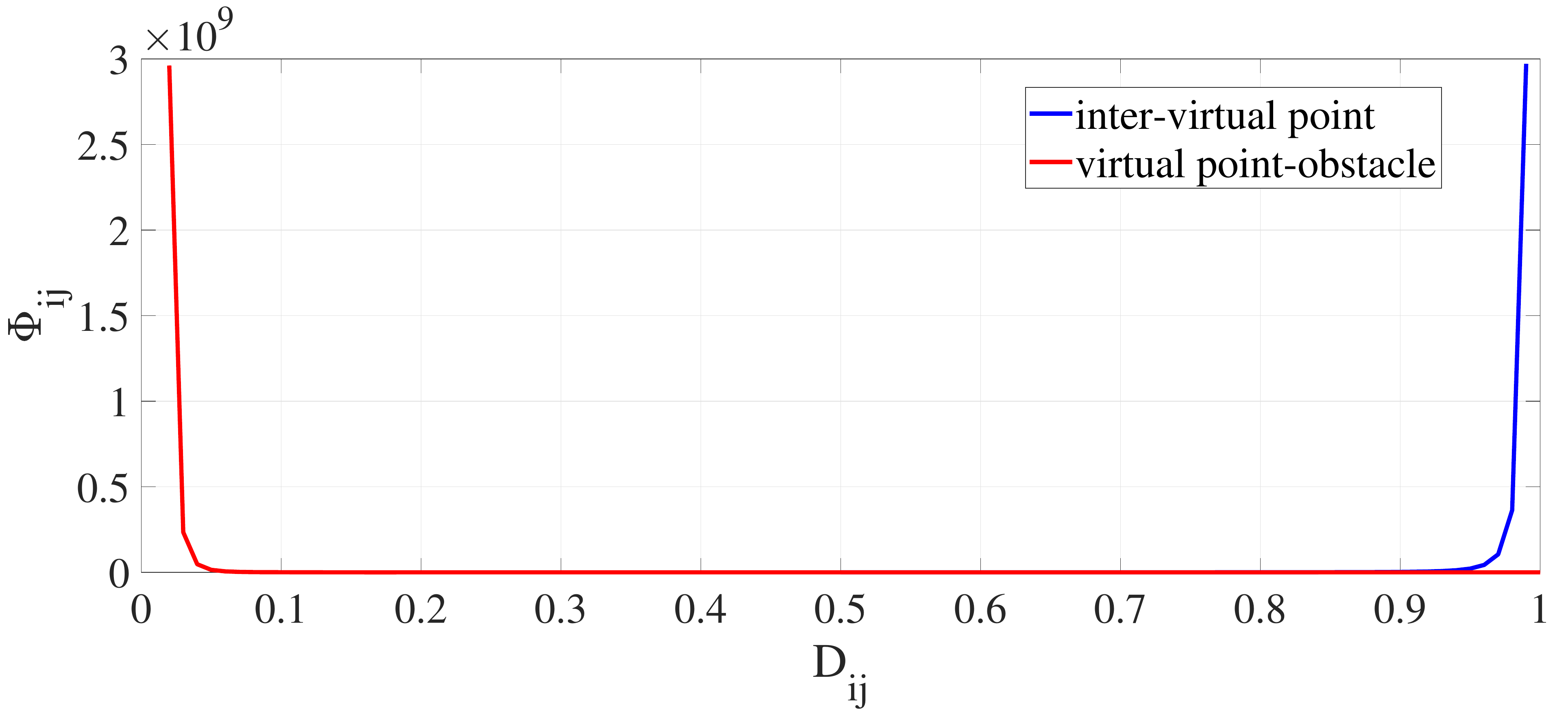}
\caption{Example of potential functions $\Psi_{ij}^f$ and $\Psi_{ij}^o$ used to avoid collisions and to maintain a desired distance.}
\label{fig:potential_function}
\end{figure}

\subsection{Distributed Swarm Control}
\label{sec:2.2}
We define the following distributed swarm control for the low-level control of heterogeneous field robots. The heterogeneous field robots consist of a group of $N$ UAVs and $N$ UGVs, and we denote $\mathcal{R}_{i}\in\Re^3$ as the position of the $i$th robot, $i=1,\dots,N$. Here, we define the virtual point (VP) $d_i:t\in\Re\mapsto p_i(t)\in\Re^3$ to be followed by $\mathcal{R}_{i}$. 

We define the dynamic undirected connectivity graph $\mathcal{C} := \{\mathcal{V}, \mathcal{E}\}$ by the vertex set $\mathcal{V} := \{1,2,\ldots,n\}$, representing the heterogeneous robots and the edge set $\mathcal{E} := \{e_{ij} : i=1,2,\ldots,n,j \in \mathcal{N}_i\}$ representing the connectivity among the heterogeneous robots, where the dynamic neighbor set $\mathcal{N}_i$ of the $i$ robot is defined as follows:
\begin{equation}
    \mathcal{N}_i := \{j \in \mathcal{V} : i~\textrm{receives information from}~j, i \neq j \}
\end{equation}

Subsequently, the kinematic evolution of VP $d_i$ generated by a distributed swarm control is as follows: For the $i$th robot,
\begin{equation}
    \dot{d_i}(t) := u^f_i+u^o_i+u^u_i,~~~~d_i(0) = \mathcal{R}_{i}(0)  \label{eq:pit}
\end{equation}
where the three control inputs $u^u_i\in\Re^3, u^f_i\in\Re^3$, and $u^o_i\in\Re^3$ represent the velocity terms. 

The first term, $u^f_i \in \Re^3$, denotes a control input to avoid a collision among heterogeneous robots, preserves connectivity, and achieves a desired formation as specified by the desired distances $\mathcal{D}_f\in\Re^+,~\forall i=1,\ldots,N, \textrm{ and } \forall j \in \mathcal{N}_i$, as defined by
\begin{equation}
    u^f_i := - \sum_{j\in\mathcal{N}_i}\frac{\partial\Phi^f_{ij}(\|d_i-d_j\|^2)^T}{\partial d_i} \label{eq:uc}
\end{equation}
where $\Phi^f_{ij}$ denotes a designed potential function to produce an attractive behavior if $\|d_i-d_j\| > \mathcal{D}_f$, a repulsive behavior if $\|d_i-d_j\| < \mathcal{D}_f$, and a null behavior if $\|d_i-d_j\| = \mathcal{D}_f$. Because there are two vertical asymptotes in $u^f_i(\Phi^f_{ij})$ that correspond to the minimum and maximum permitted distances as shown in Fig.~\ref{fig:potential_function}, collisions between robots are guaranteed and connectivity between robots is conserved.

The second term, $u^o_i\in\Re^3$, is expressed by the following equation as a control input based on a potential field that allows heterogeneous robots to avoid obstacles through a certain distance threshold $\mathcal{D}_o\in\Re^+$:
\begin{equation}
    u^o_i := -\sum_{j\in\mathcal{O}_i}\frac{\partial\Phi^o_{ij}(\|d_i-d^o_j\|)^T}{\partial d_i}  \label{eq:uo}
\end{equation}
where $\mathcal{O}_i$ denotes the set of obstacles of the $i$th VP with an obstacle point $d^o_j$ that corresponds to the position of the $r$th obstacle in the environment and $\Phi^o_{ij}$ denotes a specific artificial potential function that produces a repulsive behavior if $\|d_i-d^o_j\|<\mathcal{D}_o$ and a null behavior if $\|d_i-d^j_r\|\geq\mathcal{D}_o$. When the distance between the VP and the obstacles becomes closer to $D_o$, then the repulsive behavior increases to infinity. However, if $\|d_i-d^o_j\| \rightarrow \mathcal{D}_o$, then $\Phi_{ij}^o$ gradually converges to zero (see Fig.~\ref{fig:potential_function}).

The final term, $u^u_i \in \Re^3$, represents the desired velocity input of the VP that is controlled by the planning algorithm defined as follows:
\begin{equation}
    u^u_i = K_P(t)e_i(t)+K_I\int e_i(t)dt+K_D\frac{d}{dt}e_i(t)  \label{eq:un}
\end{equation}
where $\mathcal{T}_i\in\Re^3$ denotes the target velocity, $e_i(t)=\mathcal{T}_i-d_i$ indicates the velocity error between the target trajectory and the VP, and $K_P, K_I,$ and $K_D$ are the parameters of the desired velocity controller, respectively.

\section{DES-based High-level Control}
\label{sec:3}

\subsection{Formal Method-based System Modeling}
\label{sec:3.1}
DES is a dynamic system with characteristics of a discrete state space. It is also an event-driven system in which the state is changed by the occurred events with the passage of time. The types of discrete event models (DEMs) for modeling the DES include nondeterministic DEMs, stochastic DEMs, timed DEMs, and logical DEMs~\cite{cao1990models}. 
Basically, the state of the system and the events are listed in the order of occurrence to represent the qualitative or logical properties of the DES. Furthermore, when the next state transited by an event is uniquely determined, this system is referred to as a deterministic DES. The behavior of a deterministic DES can be expressed by listing all possible sequences from the initial state: $s=\sigma_1\sigma_2\sigma_3\cdot\cdot\cdot\sigma_i, \sigma_i\in\Sigma,i\in[1,n]$ and the initial state $x_0$. The finite sequence of events, $s$, is defined as a trace or string, and the collection of such strings is defined as a formal language. In this study, we applied automata theory with a logical DEM as the modeling formalism to describe DES behavior. 

The automaton $\mathcal{A}$ is a tuple comprising the following five elements~\cite{ramadge1989control,ramadge1987supervisory}:
\begin{equation}
    \mathcal{A}=\{\mathcal{E},\Omega,\eta,\mathcal{E}_0,\mathcal{E}_m\}
\end{equation}
where $\mathcal{E}$ is the set of states, $\Omega$ is the set of events, $\eta$ is the state transition function of $\mathcal{A}$ $(\eta:\mathcal{E} \times \Omega^*\mapsto \mathcal{E})$, $\mathcal{E}_0$ is the initial state of $\mathcal{A}$, and $\mathcal{E}_m$ is the set of marker states, which indicates a final or desired state  $(\mathcal{E}_m \subset \mathcal{E})$. In the state transition function $\eta$, $\Omega^*$ represents a sequence (string) of events comprising the null event $\varepsilon$. Moreover, the event set $\Omega$ is categorized as a set of controllable and uncontrollable events, $\Omega_c$ and $\Omega_{uc}$, respectively. The event set $\Omega$ can also be classified into observable events $\Omega_{o}$ and unobservable events $\Omega_{uo}$.

\begin{figure}[th]
\centering
\subfigure[]{\includegraphics[width=0.35\linewidth]{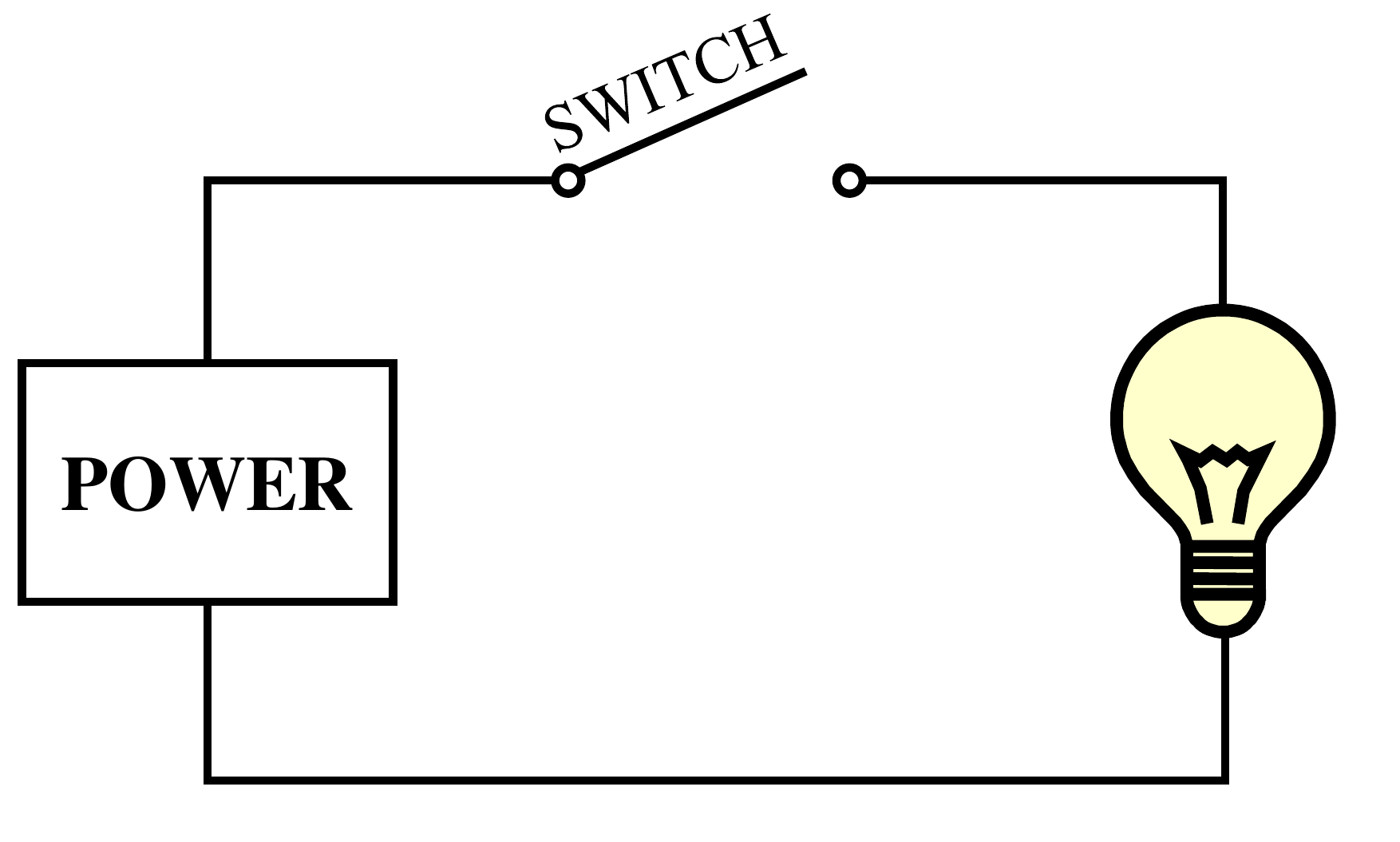}
\label{fig:ex1a}}
\hfil
\subfigure[]{\includegraphics[width=0.35\linewidth]{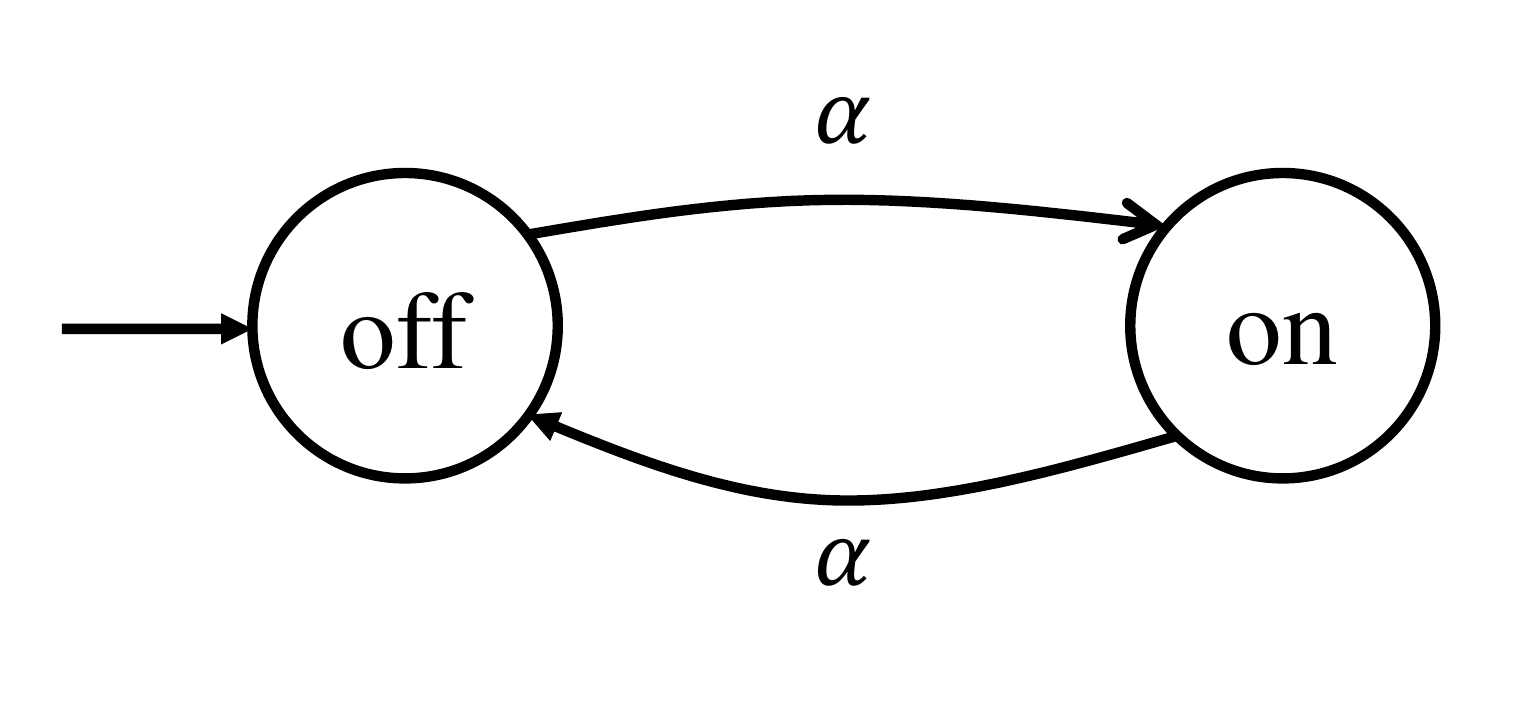}
\label{fig:ex1b}}
\caption{Example 1: (a) electrical system and (b) automata model for a light bulb.}
\label{fig:ex1}
\end{figure}

\textit{\textbf{Example 1:}} Automata modeling\\
In this example, we considered the electrical system shown in Fig.~\ref{fig:ex1}. In this system, when a control input called a flip switch event occurs, the light bulb changes state (i.e., its output). Here, automata modeling for a dynamic light bulb, as briefly depicted in Fig.~\ref{fig:ex1b}, is as follows:
$\mathcal{E} = \{\textrm{off, on}\}$, $\Omega = \{\alpha: \textrm{flip switch}\}$, $\eta = \{f(\textrm{off}, \alpha)= \textrm{on},~f(\textrm{on}, \alpha)= \textrm{off}\}$, $\mathcal{E}_0 = \textrm{off}\}$, and $\mathcal{E}_m = \{\textrm{on}\}$.

The language obtained using the automaton $\mathcal{A}$ is defined as $L(\mathcal{A}):=\{s\in\Omega^*|\eta(a_0,s)!\}$, where $\eta(a_0,s)!$ indicates that the next state in which the string $s$ occurred at $a_0$ is defined in $\mathcal{A}$. The prefix closure of language $L(\mathcal{A})$ is defined as $\overline{L(\mathcal{A})}:=\{t\in\Omega^*~|~t\le s~\exists~s\in L(\mathcal{A})\}$,
where $L(\mathcal{A})$ is defined as prefix-closed when $L(\mathcal{A})=\overline{L(\mathcal{A})}$.

The marked language of automaton $\mathcal{A}$ is defined as $L_m(\mathcal{A}):=\{s\in L(\mathcal{A})~|~\eta(a_0,s)\in A_m\}~\subseteq~L(\mathcal{A})$, where $\mathcal{A}$ satisfies $\overline{L_m(\mathcal{A})}=L(\mathcal{A})$. Then, $L(\mathcal{A})$ is defined as nonblocking. In other words, this indicates that the marked state can be attained after a string occurs in any state of $\mathcal{A}$~\cite{wonham2015supervisory}. Nonblocking should be realized for designing a proper supervisor in RW framework, because the DES may fall into deadlocks or livelocks when $L(\mathcal{A})$ is blocking.

\begin{figure}[th]
\centering
\includegraphics[trim={0.0cm 0.0cm 0.0cm 0.0cm}, width=0.4\linewidth]{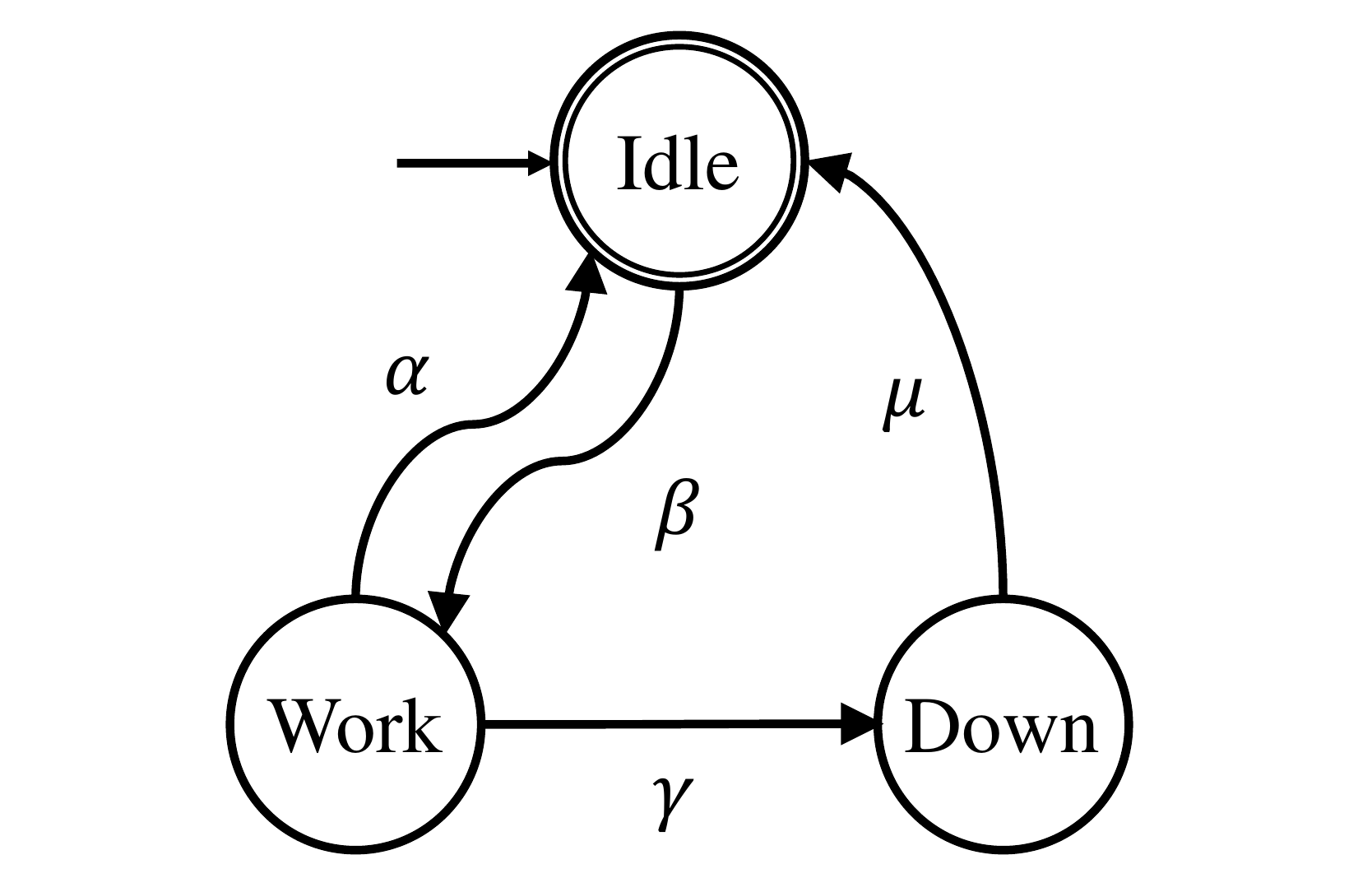}
\caption{Example 2: automata model for manufacturing machine \textbf{$\mathcal{A_{\text{mc}}}$}.}
\label{fig:ex2}
\end{figure}

\textit{\textbf{Example 2:}} Language\\
Suppose we modeled the automata model of a manufacturing machine \textbf{$\mathcal{A_{\text{mc}}}$} as shown in Fig.~\ref{fig:ex2}. When the Idle state is marked as state, the language (i.e., all strings being generable from the initial state) and marked language (i.e., all generable strings hitting some marker state) that this automaton generates are as follows: $L(\mathcal{A})=\{\varepsilon,\beta,\beta\alpha,\beta\alpha\beta,\beta\gamma\mu,\ldots\}$ and $L_m(\mathcal{A})=\{\varepsilon,\beta\alpha,\beta\gamma\mu,\ldots\}$. Here, nonblocking is satisfied when $L(\mathcal{A})=\overline{L_m(\mathcal{A})}$.

\subsection{Supervisory Control}
\label{sec:3.2}
The supervisor is defined as the automaton $\mathcal{S}=(X,\Omega,\delta,x_0,X_m)$, where $X,\Omega,\delta,x_0,$ and $X_m$ indicate the state set, event set, state transition function, initial state, and marker states, respectively. 
Consider that the plant is defined as DES $\mathcal{A}$ and that the behavior and generated language
of the plant $\mathcal{A}$ under supervision are defined as
\begin{equation}
    S/\mathcal{A}=\{X \times A, \Omega, \delta \times \eta, (x_0,a_0), X_m \times A_m\}
\end{equation}
\begin{multline}
L(S/\mathcal{A}):\;\varepsilon \in L(S/\mathcal{A}), \forall s \in \Omega^*, \varepsilon \in \Omega : \\
s \in L(S/\mathcal{A}), s\upsilon \in L(\mathcal{A}), \upsilon \not\in \Theta \Rightarrow s\upsilon \in L(S/\mathcal{A}),
\end{multline}
where $\upsilon$ is a eligible event and $\Theta$ is a control map defined as $\Theta:L(\mathcal{A}) \mapsto 2^{\Omega_c}$. 

In addition, the projection map $P:\Omega^* \mapsto \Omega_o^*$ is defined as
\begin{align}
    P(\varepsilon) & = \varepsilon \nonumber \\
    P(\upsilon) & = \left\{ \begin{array}{ll}
        \varepsilon & \textrm{if $\upsilon\not\in\Omega_o$}\\
        \upsilon & \textrm{if $\upsilon\in\Omega_o$}
        \end{array} \right. \\
    P(s\upsilon) & = P(s)P(\upsilon), \forall s \in \Omega^*, \forall \upsilon \in \Omega \nonumber
\end{align}

The controllability and observability of $L(S)$ w.r.t. $\mathcal{A}$ are described by the following definitions: 

\begin{defn}
    $\mathcal{S}$ is defined as \textit{controllable} w.r.t. $(\mathcal{A},\Omega_{uc})$ when the following condition is satisfied:
    \begin{equation}
    (\forall s,\upsilon) s \in \overline{L(S)},\; \upsilon \in \Omega_{uc},\; s\upsilon \in L(\mathcal{A}) \Rightarrow s\upsilon \in \overline{L(S)}
    \end{equation}
    \label{def:1}
\end{defn}

\begin{figure}[tb]
\centering
\includegraphics[trim={0.0cm 0.0cm 0.0cm 0.0cm}, width=0.6\linewidth]{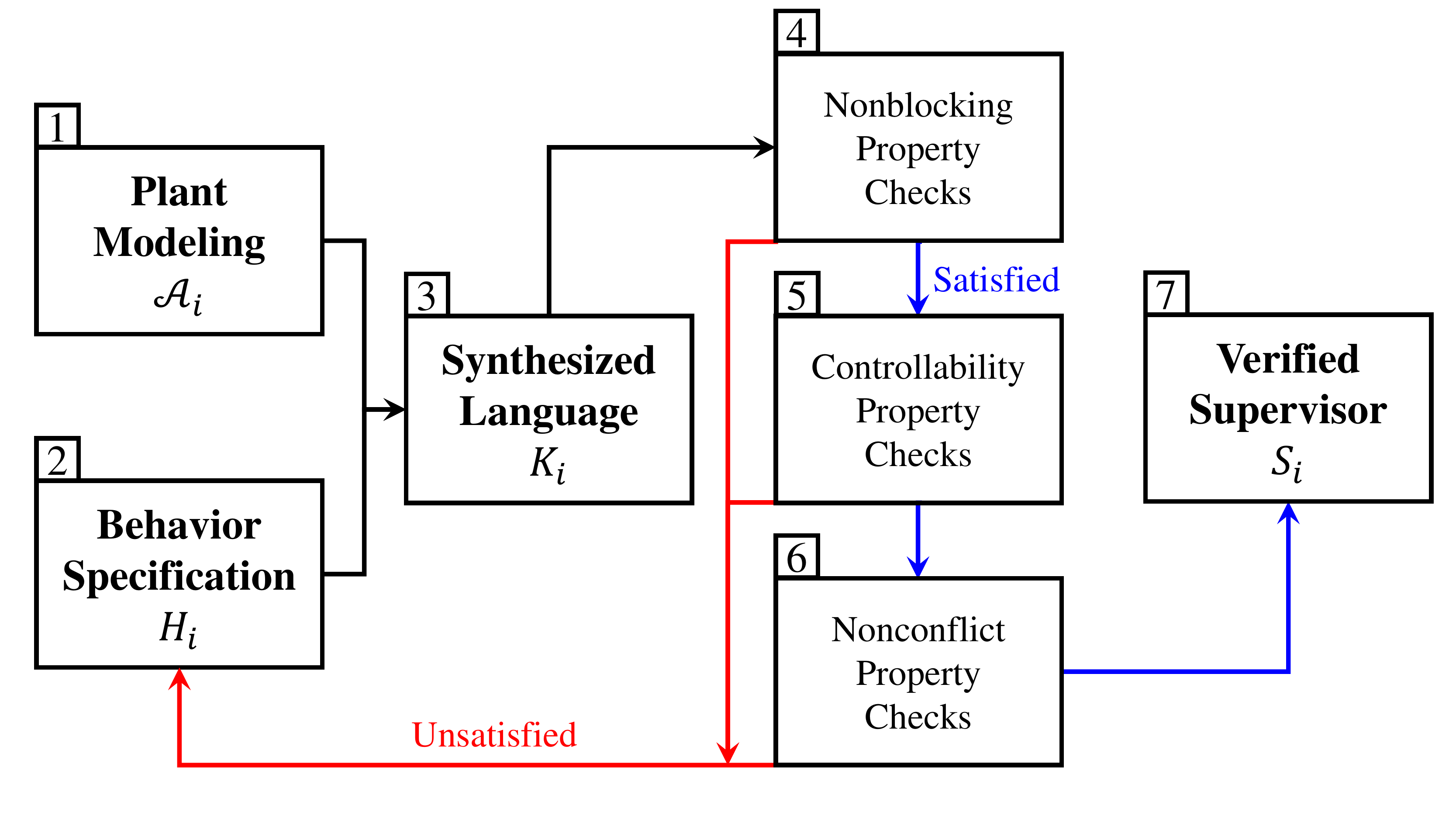}
\caption{Process of designing and verifying the supervisory controllers $\mathcal{S}_i$.}
\label{fig:sct}
\end{figure}

In other words, $s$, which is allowable by $\mathcal{S}$ and an uncontrollable event $\upsilon$, is eligible in $\mathcal{A}$, if the sequence $s\upsilon$ is eligible in $\mathcal{A}$; furthermore, if $\mathcal{S}$ also allows $s\upsilon$, then $\mathcal{S}$ is controllable w.r.t. $\mathcal{A}$.

\begin{defn}
    For $\mathcal{S} \subset \mathcal{A}$, $\mathcal{S}$ is said to be \textit{observable} \textit{w.r.t.} $(\mathcal{A},P,\Omega_{uo})$ when the following condition is satisfied:
    \begin{align}
        &(\forall s,s',\upsilon \in \overline{L(\mathcal{S})}) \upsilon \in \Omega_{uo}, P(s) = P(s'), \nonumber \\
        &s\upsilon \in \overline{L(\mathcal{S})}, s'\upsilon \in L(\mathcal{A}) \Rightarrow s'\upsilon \in \overline{L(\mathcal{S})}
    \end{align}
    \label{def:2}
\end{defn}

Namely, if the string $s\upsilon$ is permissible by the supervisor $S$ and $s'\upsilon$ is eligible in the plant $\mathcal{A}$, then $\mathcal{S}$ also has to permit $s'\upsilon$, where two strings $s,s'\in\Omega^*$ are recognized as the same string by the projection map $P$ and are also permissible by $\mathcal{S}$ and where $\upsilon$ is an unobservable event, then $\mathcal{S}$ is observable \textit{w.r.t.} $\mathcal{A}$. Note that we consider only controllability for DES-based high-level control design in this work.

Moreover, the supervisory control problem (SCP) used to design a supervisor (a procedure presented in Fig.~\ref{fig:sct}) is defined as follows:
\begin{defn}
    \label{def:3}
     For a given $K \subseteq \mathcal{A}$, we determine a supremal language $\mathcal{S}$ that is controllable w.r.t. $(\mathcal{A},\Omega_{uc})$, satisfying $L(S/\mathcal{A})=K$ and $L(S/\mathcal{A})=\overline{L_m(S/\mathcal{A})}$.
\end{defn}

Therefore, if $K$ is defined as the specification for $\mathcal{A}$, then the SCP is to identify a supervisory controller that satisfies $L(S/\mathcal{A})=K=\overline{L_m(S/\mathcal{A})}$ and is nonblocking and controllable w.r.t. $\mathcal{A}$.
Here, a plurality of supervisors that satisfy specifications exist and are controllable. Among these $K$, a supremal controllable sublanguage of $K$ is determined as the solution of the SCP. Therefore, $\mathcal{S}$ can maximally allow the eligible language to occur in $\mathcal{A}$.

\begin{figure}[th]
\centering
\subfigure[]{\includegraphics[width=0.3\linewidth]{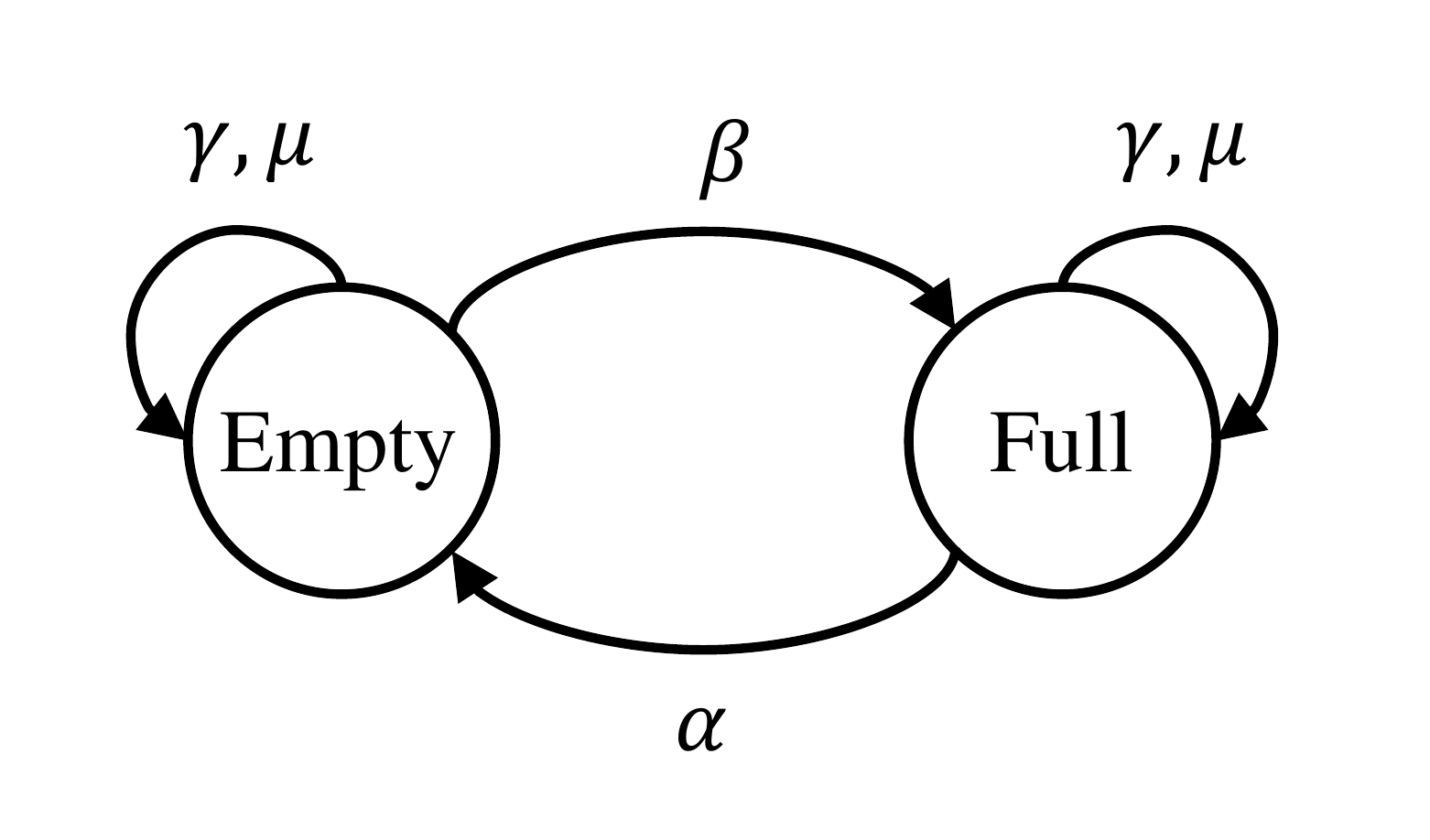}
\label{fig:ex3a}}
\hfil
\subfigure[]{\includegraphics[width=0.45\linewidth]{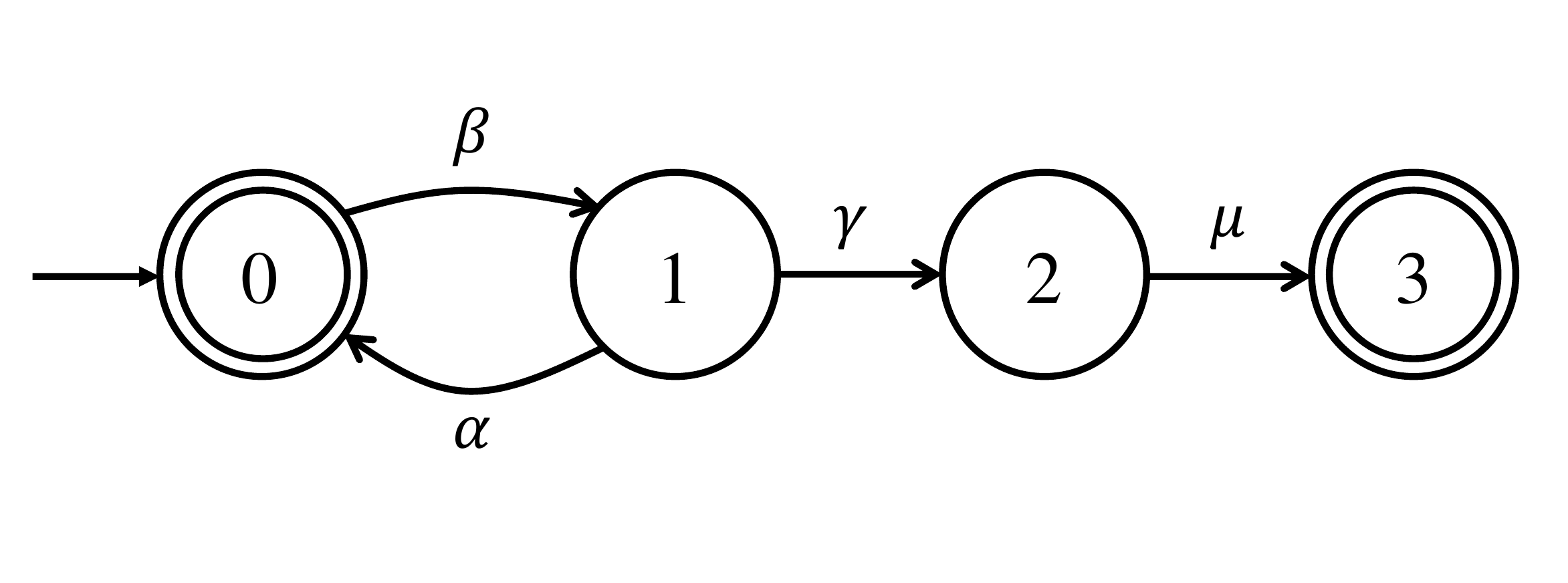}
\label{fig:ex3c}}
\hfil
\subfigure[]{\includegraphics[width=0.45\linewidth]{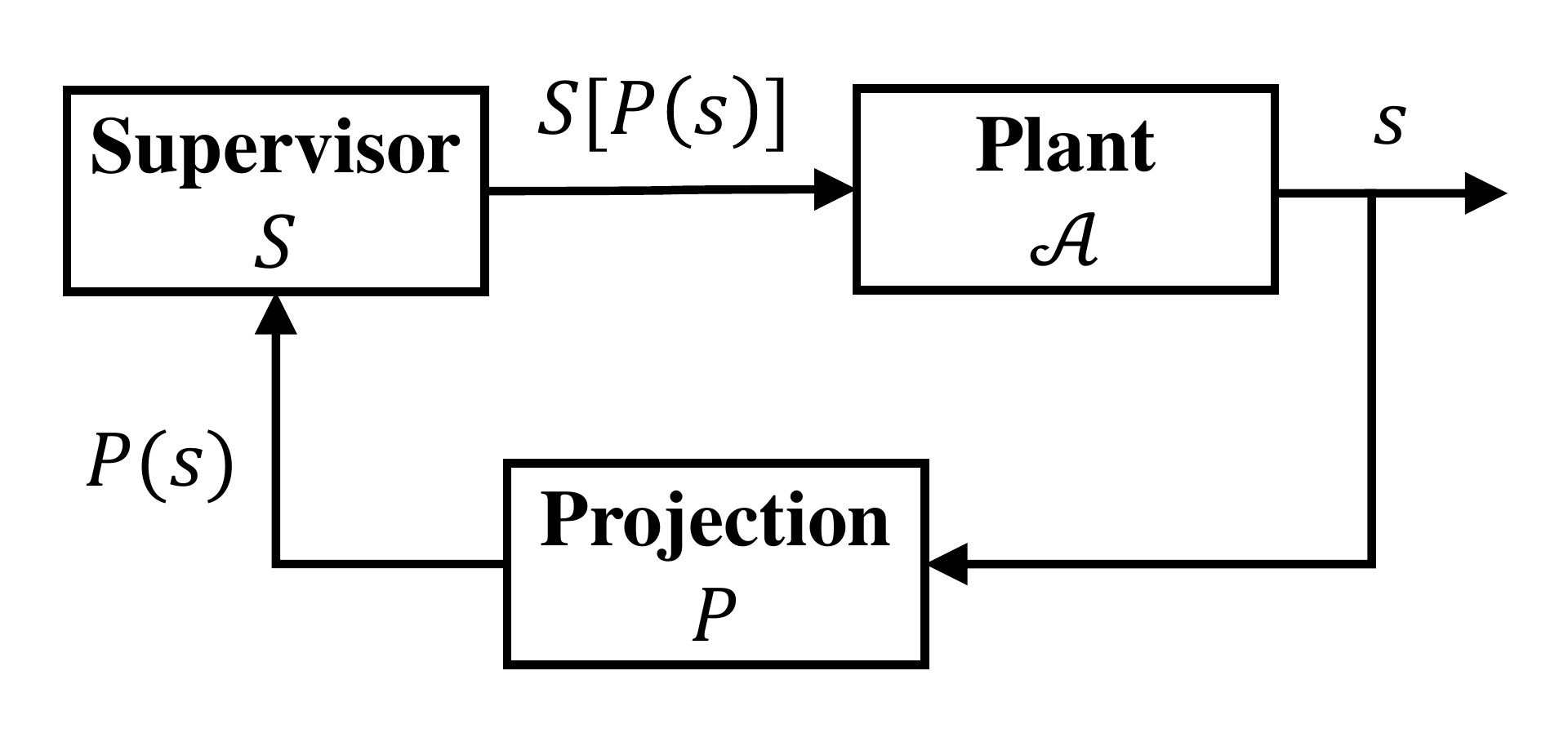}
\label{fig:ex3b}}
\caption{Example 3: (a) behavior specification $H$ for \textbf{$\mathcal{A_{\text{mc}}}$}, (b) supervisory controller $\mathcal{S}$ for $\mathcal{A}_{mc}$,  and (c) architecture of the supervisory control system.}
\label{fig:ex3}
\end{figure}

\textit{\textbf{Example 3:}} Supervisor design\\ 
When we design a manufacturing machine \textbf{$\mathcal{A_{\text{mc}}}$} with a buffer size of 2, the behavior specification can be expressed as shown in Fig.~\ref{fig:ex3a}. When defining the automaton of the specification as the alphabet $H$ and the language of $H$ as $K$ (i.e., $K=L(H)$), if $K$ satisfies the nonblocking and controllability properties concerning plant $A$, then $K$ can be defined as supervisor $\mathcal{S}$. In this case, unobservable events $\upsilon$ are reflected to $\varepsilon$ by the projection $P$, and the supervisory controller $\mathcal{S}$ controls the target plant \textbf{$\mathcal{A_{\text{mc}}}$} by control map $\Theta$. In other words, $\Theta$ associates each string $s \in L(\mathcal{A})$ with a control pattern in $\Theta(s) \in 2^{\Omega_{C}}$ marked with automata as shown in Fig.~\ref{fig:ex3c}. This approach is a closed-loop control system, the architecture of which is shown in Fig. \ref{fig:ex3b}.

\section{Hybrid Systems-based Hierarchical Control Architecture}
\label{sec:4}

\subsection{Hybrid Automata}
\label{sec:4.1}
Hybrid systems are dynamic systems that combine CTS and DES. For example, a motor (a continuous component) of a mobile robot controlled by a microprocessor (a discrete component) is a hybrid system. There are various hybrid systems deployed in real applications (e.g., elevators, automobile engines, heating systems, and transportation systems). For this purpose, hybrid automatons have been proposed as formal models for hybrid systems~\cite{henzinger2000theory}. The hybrid automaton $\mathcal{G}_h$ is a tuple comprising the following elements: 
\begin{equation}
    \mathcal{G}_h=(\mathcal{E},\mathcal{X},\mathcal{E},\mathcal{U},\mathcal{F},\phi,Inv,Guard,\rho,\mathcal{E}_0,\mathcal{X}_0)    
\end{equation}
where $\mathcal{E}$ is the set of discrete states, $\mathcal{X}$ is the set of continuous states, $\mathcal{E}$ is the set of events, $\mathcal{U}$ is the set of admissible controls, $\mathcal{F}$ is the vector field of $\mathcal{G}_h$ ($\mathcal{F}: \mathcal{E} \times \mathcal{X} \times \mathcal{U} \rightarrow \mathcal{X}$), $\phi$ is the discrete state transition function of $\mathcal{G}_h$ ($\phi: \mathcal{E} \times \mathcal{X} \times \mathcal{E} \rightarrow \mathcal{E}$), $Inv$ is the set defining an invariant condition ($Inv \subseteq \mathcal{E} \times \mathcal{X}$), $Guard$ is the set defining a guard condition ($Guard \subseteq \mathcal{E} \times \mathcal{E} \times \mathcal{X}$), $\rho$ is the reset function ($\rho: \mathcal{E} \times \mathcal{E} \times \mathcal{X} \times \mathcal{E} \rightarrow \mathcal{X}$), $\mathcal{X}_0$ is the initial discrete state, and $\mathcal{X}_0$ is the initial continuous state. In this study, we also attempt to model heterogeneous field robots using hybrid automata models that combine continuous time models and DEMs. Furthermore, we extend the SCT introduced in Section~\ref{sec:3} to propose an HSHC structure applicable to complex dynamic systems. The detailed system modeling and supervisor design are discussed in the following sections.

\begin{figure}[t]
\centering
\includegraphics[trim={0.0cm 0.0cm 0.0cm 0.0cm}, width=0.6\linewidth]{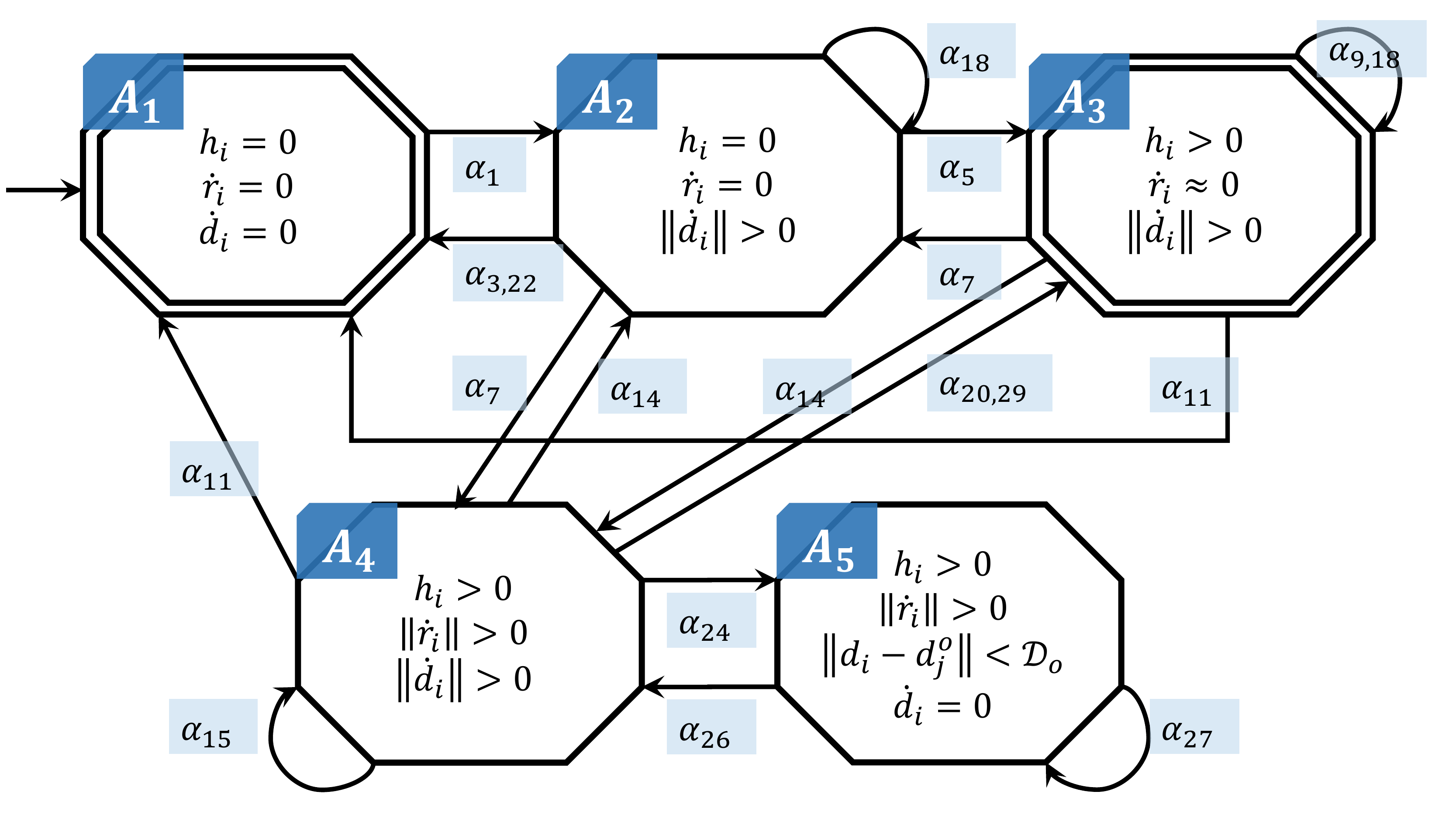}
\caption{Hybrid automata model for UAV $\mathcal{G_A}$.}
\label{fig:uav}
\end{figure}

\begin{figure}[tb]
\centering
\includegraphics[trim={0.0cm 0.0cm 0.0cm 0.0cm}, width=0.6\linewidth]{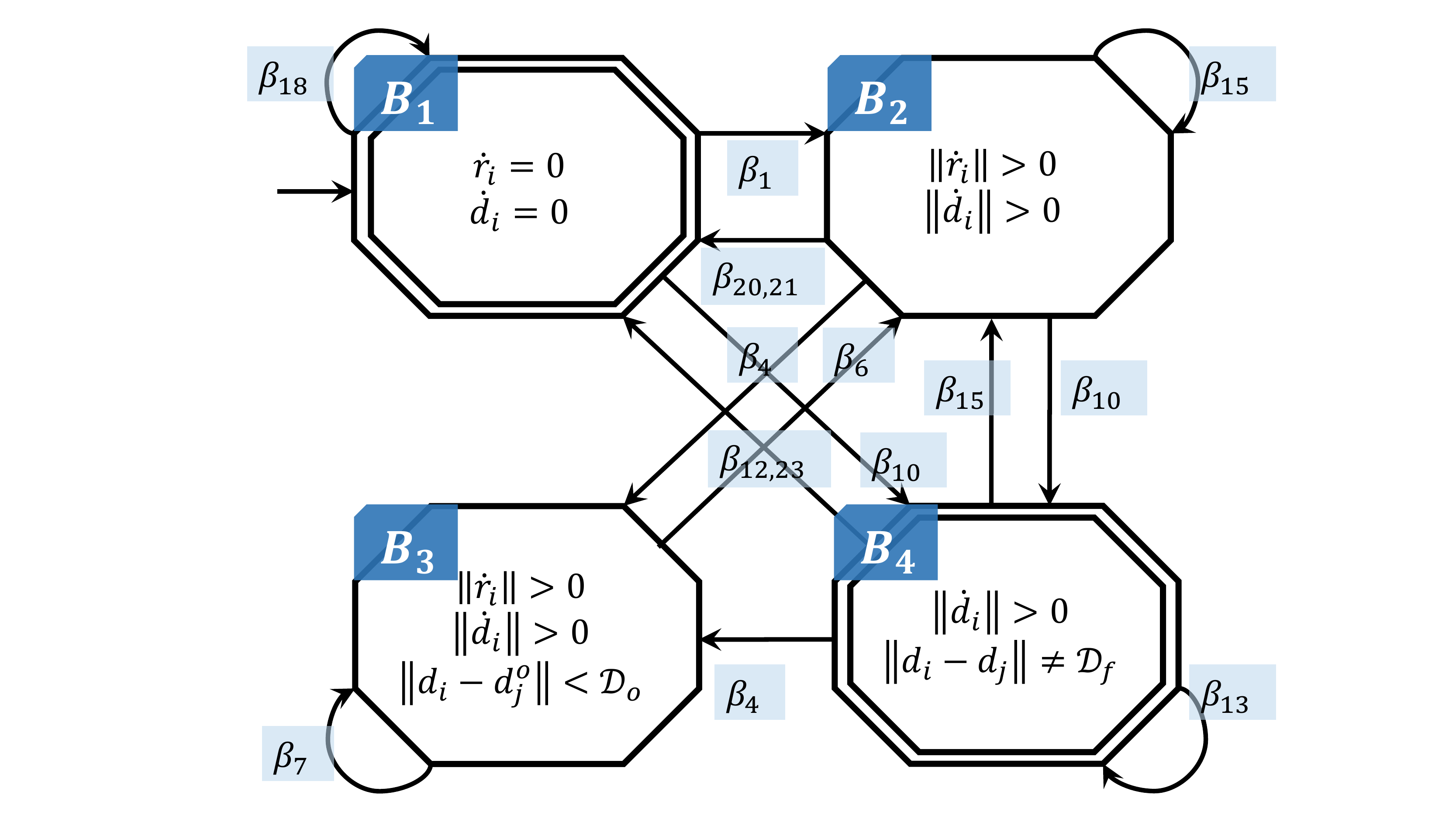}
\caption{Hybrid automata model for UGV $\mathcal{G_B}$.}
\label{fig:ugv}
\end{figure}

\begin{table}[tbp]
\caption{Description of events for hybrid UAV model $\mathcal{G_A}$ and hybrid UGV model $\mathcal{G_B}$}
\label{table_ev}
\centering
\begin{tabular}{cccl}
\hline 
\hline
Plant                                                                   & Event       & Controllability & \multicolumn{1}{c}{Description} \\ \hline \hline
\multirow{15}{*}{\begin{tabular}[c]{@{}c@{}}UAV\\ ($\mathcal{G_A}$)\end{tabular}} & $\alpha_1$  & O               & Arm                             \\
                                                                        & $\alpha_3$  & O               & Disarm                          \\
                                                                        & $\alpha_5$  & O               & Take off                        \\
                                                                        & $\alpha_7$  & O               & Land                            \\
                                                                        & $\alpha_9$  & O               & Keep hovering                   \\
                                                                        & $\alpha_{11}$ & O               & Return to home                  \\
                                                                        & $\alpha_{14}$ & X               & Start mission                   \\
                                                                        & $\alpha_{15}$ & O               & Keep mission                    \\
                                                                        & $\alpha_{18}$ & X               & Receive mission                 \\
                                                                        & $\alpha_{20}$ & X               & Finish mission                  \\
                                                                        & $\alpha_{22}$ & X               & Time out                        \\
                                                                        & $\alpha_{24}$ & X               & Detect obstacles                \\
                                                                        & $\alpha_{26}$ & X               & Detect free space                \\
                                                                        & $\alpha_{27}$ & O               & Keep avoiding                   \\
                                                                        & $\alpha_{29}$ & X               & Clear mission                   \\ \hline \hline
\multirow{12}{*}{\begin{tabular}[c]{@{}c@{}}UGV\\ ($\mathcal{G_B}$)\end{tabular}} & $\beta_1$   & O               & Start mission                   \\
                                                                        & $\beta_4$   & X               & Detect obstacles                \\
                                                                        & $\beta_6$   & X               & Detect free space                \\
                                                                        & $\beta_7$   & O               & Keep avoiding                   \\
                                                                        & $\beta_{10}$  & X               & Network connected               \\
                                                                        & $\beta_{12}$  & X               & Network disconnected            \\
                                                                        & $\beta_{13}$  & O               & Keep formation                  \\
                                                                        & $\beta_{15}$  & O               & Keep mission                    \\
                                                                        & $\beta_{18}$  & X               & Receive mission                 \\
                                                                        & $\beta_{20}$  & X               & Finish mission                  \\
                                                                        & $\beta_{21}$  & O               & Clear mission                   \\
                                                                        & $\beta_{23}$  & O               & Break formation                 \\ \hline
                                                                        \hline
\end{tabular}
\end{table}

\subsection{Plant Modeling for Heterogeneous Field Robots}
\label{sec:4.2}
The objective is to enable collaboration among heterogeneous field robots, so that the hybrid automaton model and behavior specifications of heterogeneous field robots consisting of UAVs and UGVs can be designed. Hybrid models for UAVs and UGVs are shown in Figs. \ref{fig:uav} and \ref{fig:ugv}, respectively. In these figures, the initial state is shown first (i.e., at the upper right position), the marked states are indicated by a double line, the discrete states are listed at the top left of the shape, the continuous states are described in the middle of the shape, and the events and state transition functions are specified with alphabetical designations and arrows, respectively. Here, we modeled the states of hybrid models by focusing on the dynamics of the heterogeneous robots with a CTS, unlike in previous studies~\cite{8834867}. In other words, hybrid modeling was considered for heterogeneous field robots by combining the CTS-based modeling introduced in Section~\ref{sec:2} and the DES-based modeling introduced in Section~\ref{sec:3}. As a result, UAV (model as $\mathcal{G_A}$) consists of 5 states, 23 transitions, and 15 events whereas UGVs (model as $\mathcal{G_B}$) consist of 4 states, 13 transitions, and 12 events. The events for each model are described in detail in Table~\ref{table_ev}. Here, odd-numbered events are controllable events and even-numbered events are uncontrollable events. Finally, we obtained the hybrid automata model of the entire plant system through parallel composition~\cite{wonham2015supervisory} as follows:
\begin{equation}
    \mathcal{G}_{plant} = \mathcal{G_A} || \mathcal{G_B}^1 || \mathcal{G_B}^2
\end{equation}

\subsubsection{Hybrid Automata Model for UAVs}
Hybrid automata $\mathcal{G_A}$ models (shown in Fig.~\ref{fig:uav}) the dynamic behavior of the UAV, and the states are as follows. $\mathcal{Q_A}=\{A_1,A_2,A_3,A_4,A_5\}$, where $A_1$: \textit{Ideal}, $A_2$: \textit{Arming}, $A_3$: \textit{Hovering}, $A_4$: \textit{Flying}, and $A_5$: \textit{Avoiding}. The eligible events $\mathcal{E_A}$ are detailed in Table~\ref{table_ev}, and the discrete transition function is $\phi_A: f(\mathcal{G_A},\mathcal{E_A})=\mathcal{G_A}$, as shown by the arrow in Fig.~\ref{fig:uav}. The initial state $\mathcal{E}_0$ is $A_1$ and the desired state is $\mathcal{E}_m=\{A_1,A_3\}$. From the perspective of the CTS, $h_i$ represents the height of the $i$th robot, $\mathcal{R}_{i}$ is the position, and $d_i$ is the position of the $i$th VP, as defined in Section~\ref{sec:2}. For example, if an event $\alpha_5$ occurs in state $A_2$, the state transitions to $A_3$, where the continuous state $\mathcal{X_A}$ is a state where the height $h_i$ of the UAVs is greater than zero, the velocity $\mathcal{R}_{i}$ is close to zero, and a control command is input to the VP ($||d_i|| > 0$). The initial continuous state of this model is as follows: $h_i, \dot{\mathcal{R}_{i}}, \dot{d_i} = 0$

\subsubsection{Hybrid Automata Model for UGVs}
Hybrid automata $\mathcal{G_B}$ also model (shown in Fig.~\ref{fig:ugv}) the dynamic property of the UGV, and the states are as follows. $\mathcal{Q_B}=\{B_1,B_2,B_3,B_4\}$, where $B_1$: \textit{Stationary}, $B_2$: \textit{Navigation}, $B_3$: \textit{Safety}, and $B_4$: \textit{Formation}. The discrete transition function is $\phi_B: f(\mathcal{G_B},\mathcal{E_B})=\mathcal{G_B}$, as shown by the arrow in Fig.~\ref{fig:uav}, the initial state $\mathcal{E}_0$ is $B_1$, and the desired state is $\mathcal{E}_m=\{B_1,B_4\}$. For example, when UGVs form the desired formation (i.e., the state is $B_4$) and an eligible event $\beta_4$ occurs that detects an obstacle, the state transitions to $B_3$. In the $B_3$ state, the event of avoiding obstacles ($\beta_7$) is allowed by the supervisory controller $\mathcal{S}_i$. Here, in the CTS, the velocity of the UGV ($||\dot{\mathcal{R}_{i}}||$) and the control input to the VP ($||\dot{d_i}||$) are greater than zero, and the distance between the UGV and the obstacle ($||d_i-d_j^o||$) is less than the desired safe distance ($\mathcal{D}_o$).

\begin{figure*}[tp]
    \centering
    \subfigure[]
    {
        \includegraphics[trim={2cm 0 2cm 0},width=0.3\linewidth,page=1]{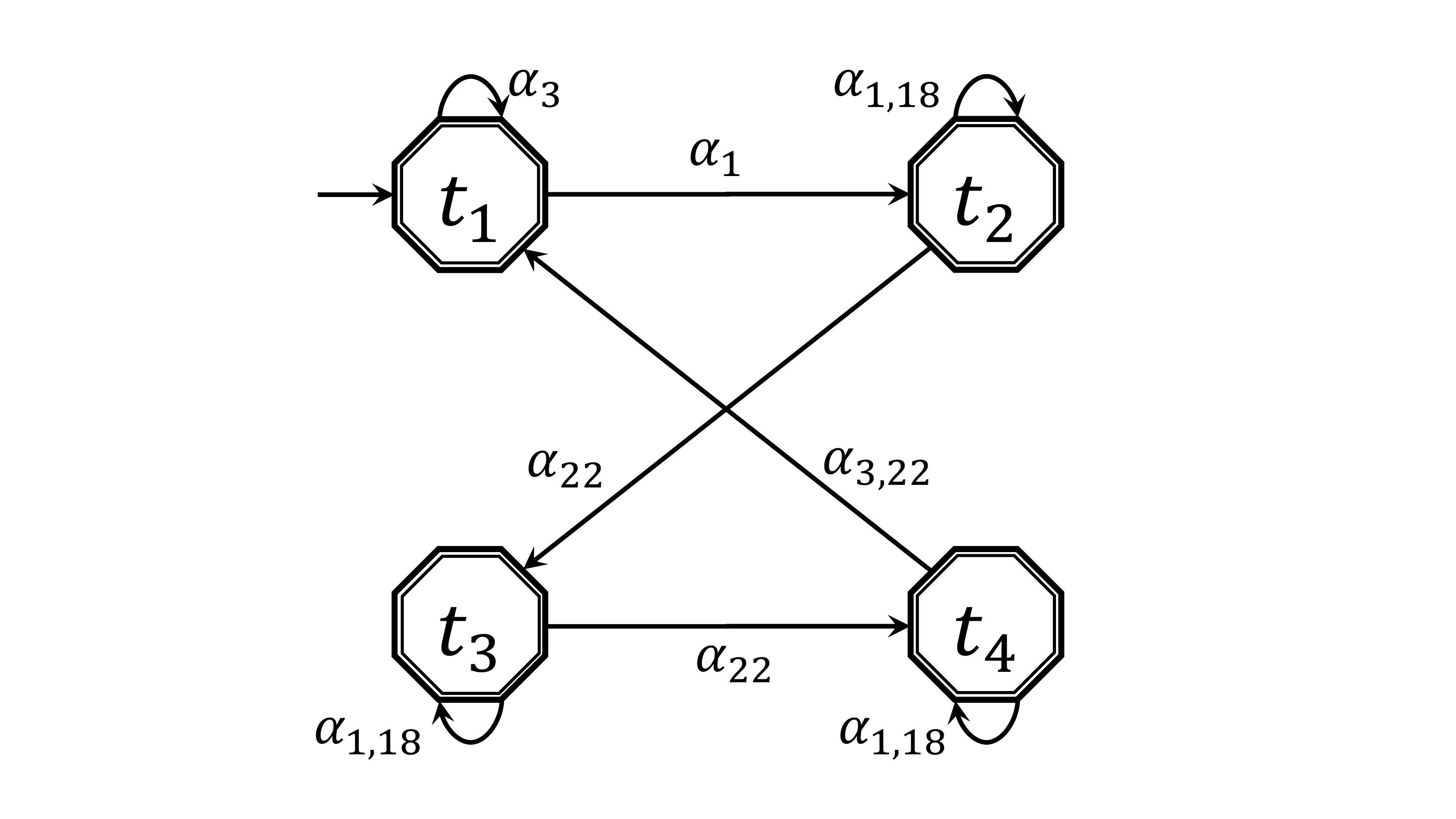}
        \label{fig:spec_1}
    }
    \subfigure[]
    {
        \includegraphics[trim={2cm 0 2cm 0},width=0.3\linewidth,page=2]{spec.pdf}
        \label{fig:spec_2}
    }
    \subfigure[]
    {
        \includegraphics[trim={2cm 0 2cm 0},width=0.3\linewidth,page=3]{spec.pdf}
        \label{fig:spec_3}
    }
    \\
    \subfigure[]
    {
        \includegraphics[trim={2cm 0 2cm 0},width=0.3\linewidth,page=4]{spec.pdf}
        \label{fig:spec_4}
    }
     \subfigure[]
    {
        \includegraphics[trim={2cm 0 2cm 0},width=0.3\linewidth,page=5]{spec.pdf}
        \label{fig:spec_5}
    }
    \subfigure[]
    {
        \includegraphics[trim={2cm 0 2cm 0},width=0.3\linewidth,page=6]{spec.pdf}
        \label{fig:spec_6}
    }
\caption{Specifications for heterogeneous robots: (a) flying for UAV $H^1_A$, (b) obstacle avoidance for UAV $H^2_A$, (c) mission management for UAV $H^3_A$, (d) navigation for UGV $H^1_B,$ (e) obstacle avoidance for UGV $H^2_B$, and (f) mission management for UAV $H^3_B$.}
\label{fig:specifications}
\end{figure*}

\subsection{Design of Specifications}
\label{sec:4.3}
The supervisor is the result of establishing a legal language (i.e., specification) and then finding a supremal controllable sublanguage for the plant. Alternatively, we should find a controllable language that meets the conditions for becoming an appropriate supervisor w.r.t. plant model $\mathcal{G}_i$. The behavior specifications are designed such that they satisfy the condition of nonblocking, controllability, and nonconflictness for cooperation of heterogeneous field robots. The $POL_1, POL_2, \ldots, POL_{10}$ policies for designing specifications are as follows:

\begin{itemize}
    \item $POL_{1}$: Arming must be in operation before the UAV takes off (i.e., hovering).
    \item $POL_{2}$: The mission is assigned while the UAV is arming and performed while the UAV is hovering.
    \item $POL_{3}$: When UGV completes its mission, the UAV lands and waits for the next mission.
    \item $POL_{4}$: UGVs form a desired formation when the network is connected.
    \item $POL_{5}$: UGVs receive a mission and start while the UGV is stop.
    \item $POL_{6}$: UGVs perform missions forming formations.
    \item $POL_{7}$: At the end of the mission, the UAV remains hovering and the UGV remains stopped.
    \item $POL_{8}$: If an obstacle is found while the robot is moving, obstacle avoidance is the priority and guaranteed.
    \item $POL_{9}$: Duplicate missions are not assigned to each robot.
    \item $POL_{10}$: When the mission is over, wait for the next mission.
\end{itemize}

Therefore, we modeled the specifications $H_i$ that met the above-mentioned policies before synthesizing the modular supervisory controller $\mathcal{S}_i$. A total of six specifications were designed to achieve the desired control objectives for each robot, as shown in Fig.~\ref{fig:specifications}. Figs.~\ref{fig:specifications}a, \ref{fig:specifications}b, and \ref{fig:specifications}c are the control specifications for the UAV, modeling for arming and disarming, obstacle avoidance, and mission management, respectively. Additionally, Figs. \ref{fig:specifications}d, \ref{fig:specifications}e, and \ref{fig:specifications}f are control specifications for navigation, obstacle avoidance, and mission management for UGVs. 
In details, we build automaton $H_A^1$ shown in Fig.~\ref{fig:spec_1}, that take into account for policies $POL_1$, $POL_2$, and $POL_7$.  The states of $H_A^1$ are as follows: (1) state $t_1$ represents the UAV is stopped, (2) state $t_2$ represents the arming state where the UAV operates and is assigned a desired goal, (3) the UAV is driven along with a given mission in $t_3$, and (4) the state $t_4$ shows the situation in which the UAV avoids obstacles. We also built an automaton $H_B^1$, as shown in Fig.~\ref{fig:spec_4}, for the specifications of the UGV, taking into account $POL_3$ to $POL_7$. Specifications $ H_A ^ 2 $ and $ H_B ^ 2 $ considering $ POL_8 $ are shown in Figs.~\ref{fig:spec_2} and~\ref{fig:spec_5}. In other words, if obstacles are found in states $ o_1 $ and $ v_1 $ while the robot is in motion, the states will be transitioned to states $ o_2 $ and $ v_2 $, respectively. Then, the obstacle avoidance controls, controllable events, $ \ alpha_{27} $ and $ \ beta_7 $ will be allowed in that states. While heterogeneous robots avoid obstacles, the state transitions from $ o_2 $ to $ o_3 $ and $ v_2 $ to $ v_3 $ are repeated before free space detect events $\alpha_{26}$ and $\beta_6$ occur. Finally, specifications $H_A^3$ and $H_B^3$ reflecting $ POL_9 $ and $ POL_{10} $ are depicted in Figs.~\ref{fig:spec_3}  and~\ref{fig:spec_6}. In the case of UGVs, the state $s_1$ that receives the mission transitions to the state $s_2$ when the mission begins. In the state $s_2$, the mission is performed according to the controllable event $\beta_{15}$. Here, state $s_2$ transitions to state $s_4$ when the mission is completed, and transitions to state $s_3$ when the mission is cleared. In the states $s_3$ and $s_4$, when a new mission is received, the state transitions to $s_1$ and iterates through these mission loop.

Therefore, the control objectives model what events would be allowed and disallowed correctly by the supervisor when heterogeneous robots do not collide with each other when given a mission. In addition, when the obstacle is found, it gives the highest priority; in case of UGVs, it is designed to keep the formation while following the given path. The design specifications were modeled to reflect the desired policies and meet the conditions for the proper formal language $K_i$.

\begin{algorithm}[tb]
\caption{Solutions of modular supervisory SCP}
\label{al:scp}
\begin{algorithmic}
\STATE \textbf{Input:} Plant $\mathcal{G}_i$ and specification $K_i,~i=1,2,\dots,n$.
\STATE \textbf{Output:} Modular supervisors $\mathcal{S}_j$ or $\mathcal{S}’_j,~j=1,2,\dots,m$.\\
\begin{itemize}
    \item Step 1: Define the subplant automaton $\mathcal{G}_{sub,i}$ and specification automaton $K_i$.
    \item Step 2: Determine $\mathcal{S}_j$ (the supremal controllable sublanguage) of $K_i$ and $\mathcal{G}_{sub,i}$.
    \item Step 3: Verify the controllability of $K_i$ w.r.t. $\mathcal{G}_{sub,i}$ using $\Omega_c$ in $K_i$.
    \item Step 4: Verify the nonblocking of $K_i$.
    \item Step 5: Verify the nonconflicting of $K_i$ w.r.t. $\mathcal{G}_{sub,i}$. If $K_i$ is nonconflicting, then $\mathcal{S}’_j$ exist, where $\mathcal{S}’_j = K_i$. Goto Step 8.
    \item Step 6: If the specification $K_i$ does not satisfy steps 3, 4 and 5, go back to step 1 and redesign. Nevertheless, if $K_i$ does not meet the proper conditions then, $\mathcal{S}’_j$ does not exist. Goto Step 7.
    \item Step 7: $\mathcal{S}_j$ is the solution of the SCP w.r.t. ($K_i,G_{sub,i}$). \textbf{return} $\mathcal{S}_j$

    \item Step 8: If $L(S_j)=L(S’_j/G_{sub,i})$, then $\mathcal{S}’_j$ is the solution of the SCP w.r.t. ($K_i,G_{sub,i}$). \vfil \textbf{return} $\mathcal{S}'_j$, \textbf{else return} $\mathcal{S}_j$. 
\end{itemize}
\end{algorithmic}
\end{algorithm}

\begin{figure*}[tp]
    \centering
    \subfigure[]
    {
        \includegraphics[trim={4cm 0 4cm 0},width=0.8\linewidth,page=1]{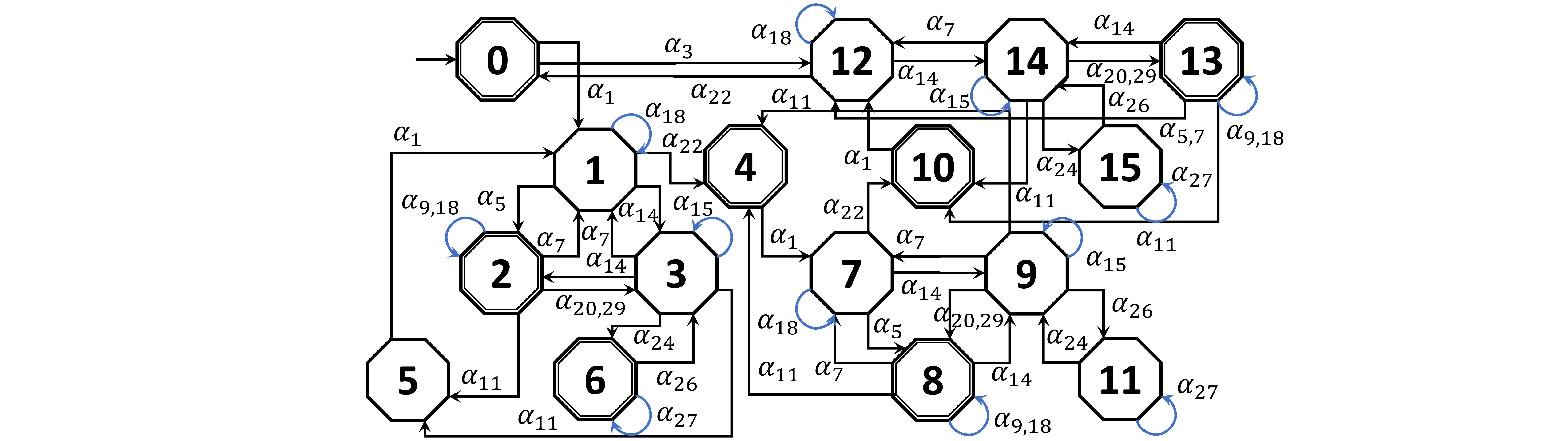}
        \label{fig:super_1}
    }
    \\
    \subfigure[]
    {
        \includegraphics[trim={4cm 0 4cm 0},width=0.8\linewidth,page=2]{supervisor.pdf}
        \label{fig:super_2}
    }
    \caption{UAV modular supervisors ($\mathcal{S}_\mathcal{A}^1$ and $\mathcal{S}_\mathcal{A}^3$).}
\label{fig:uav_supervisor}
\end{figure*}

\begin{figure}[tp]
    \centering
        \includegraphics[trim={2cm 0 2cm 0},width=0.4\linewidth,page=1]{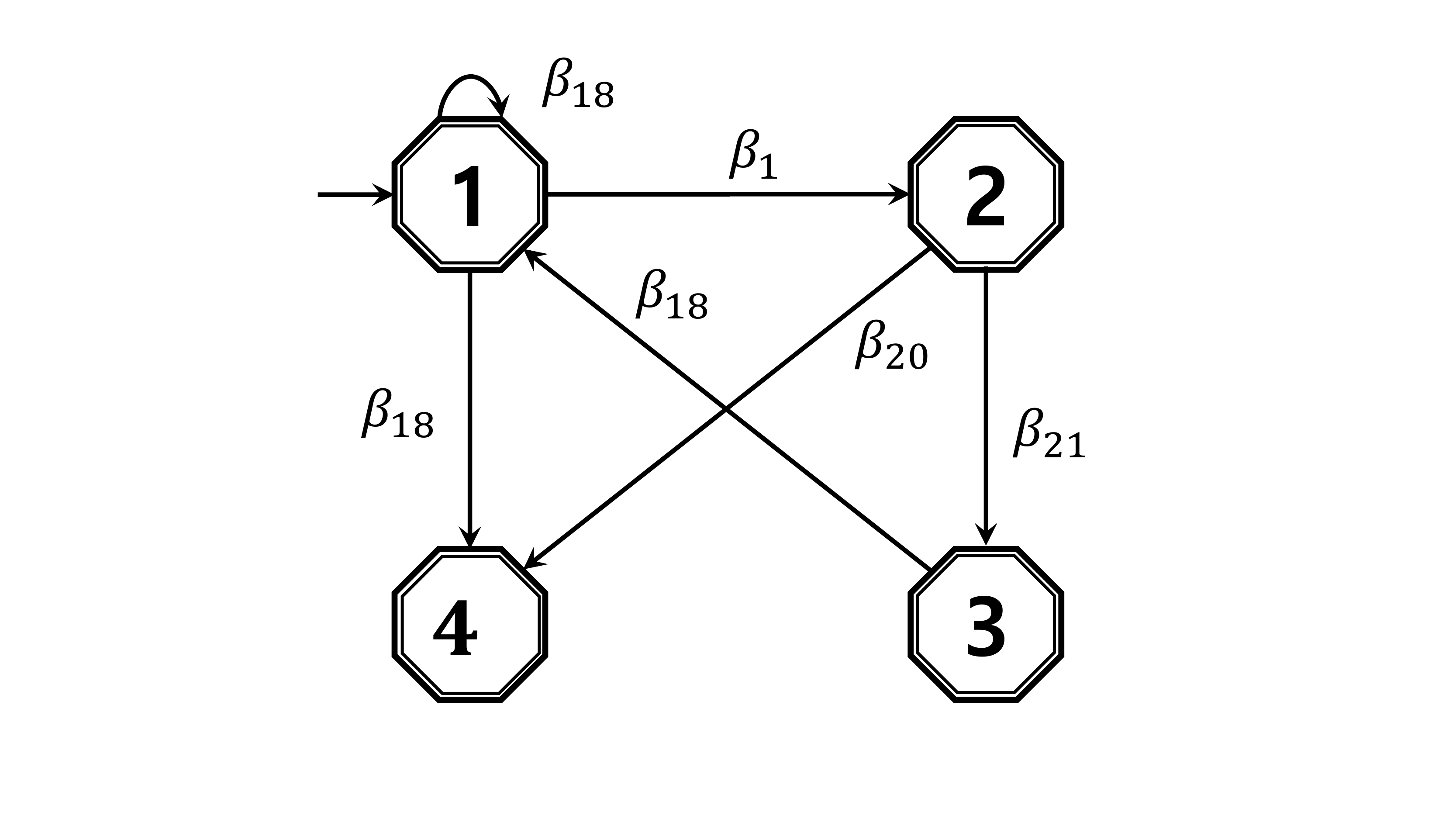}
        \label{fig:super_3}
    \caption{UGV modular supervisor ($\mathcal{S}_\mathcal{B}^3$).}
\label{fig:ugv_supervisor}
\end{figure}

\subsection{Synthesis of the Modular Supervisor}
\label{sec:4.4}
For modular supervisory control, plant $\mathcal{G}_i$ and specification $H_i$ are synthesized to obtain the legal language $K_i$. If $\mathcal{S}_j,~j=1,2,\dots,m$, is verified for such legal specifications $K_i,~i=1,2,\dots,n$ ($m\leq n$), then this $\mathcal{S}_j$ is defined as a modular supervisory controller. The modular supervisor $\mathcal{S}_j$ satisfies the nonconflictness condition defined as
\begin{equation}
    \bar{S} = \bar{S_1} \land \bar{S_2} \land \dots \land \bar{S_m}
    \label{eq:central_s}
\end{equation}
where $\land$ is meet product~\cite{wonham2015supervisory} (defined as $S_1 \land S_2 = L(S_1) \bigcap L(S_2)$). In (\ref{eq:central_s}), $S$ is a centralized supervisor designed w.r.t. $K = K_1 \land \ldots \land K_n$. If $\mathcal{S}_j$ satisfies nonconflictness, all modular supervisors can control the plant $\mathcal{G}_{plant}$ like the centralized supervisor $\mathcal{S}$. If $\mathcal{S}_j$ is conflicting, modular supervisors do not satisfy the specifications $K_i,~i=1,2,\dots,n$ by allowing unnecessary events. Therefore, the modular supervisory $\mathcal{S}_j$ must satisfy the conditions of controllability, nonblocking, and nonconflictness as explained in Section~\ref{sec:3}. The solution of the modular supervisory SCP is presented in Algorithm~\ref{al:scp} (see~\cite{son2011design} for a formal proof).

Subcomponent $\mathcal{G}_{sub,i}$ of the entire plant $\mathcal{G}_{plant}$ consists of UAV $\mathcal{G_A}$ and UGV $\mathcal{G_B}$. First, we check whether each specification $K_i$ satisfies controllability, nonblocking, and nonconflictness for the plant. According to Algorithm~\ref{al:scp}, the specifications $K_i$ that meet the requirements for  subplants UAV $\mathcal{G_A}$ and UGV $\mathcal{G_B}$, $K_\mathcal{A}^3$, $K_\mathcal{B}^1$, and $K_\mathcal{B}^2$, are defined as modular supervisors $\mathcal{S}_\mathcal{A}^3$, $\mathcal{S}_\mathcal{B}^1$, and $\mathcal{S}_\mathcal{B}^2$, respectively. In contrast, the supremal controllable sublanguage $\mathcal{S}_j$ was synthesized for the specifications $K_i$ that do not satisfy the controllability, that is, $K_\mathcal{A}^1$, $K_\mathcal{A}^2$, and $K_\mathcal{B}^3$. The obtained supremal controllable sublanguage $\mathcal{S}_j$ is the solution to the SCP for $K_i$ and $G_i$; hereafter, it is determined to be supervisors $\mathcal{S}_\mathcal{A}^1$, $\mathcal{S}_\mathcal{A}^2$, and $\mathcal{S}_\mathcal{B}^3$, respectively, as shown in Figs.~\ref{fig:uav_supervisor} and~\ref{fig:ugv_supervisor}. As a result, we found a modular supervisor $\mathcal{S}_j,~j=1,2,\ldots,6$, for specification language $K_i$ w.r.t. $\mathcal{G}_i$. Finally, we verified that the designed modular supervisory controllers satisfy nonconflictness by using a supervisory control synthesis tool. These modular supervisors $\mathcal{S}_j$ can control the dynamic complex system in the desired behavior by selectively allowing or disallowing controllable events $\mathcal{E}$ generated by the plant $\mathcal{G}_{plant}$.

\begin{figure}[tb]
    \centering
    \includegraphics[width=0.55\linewidth,page=1]{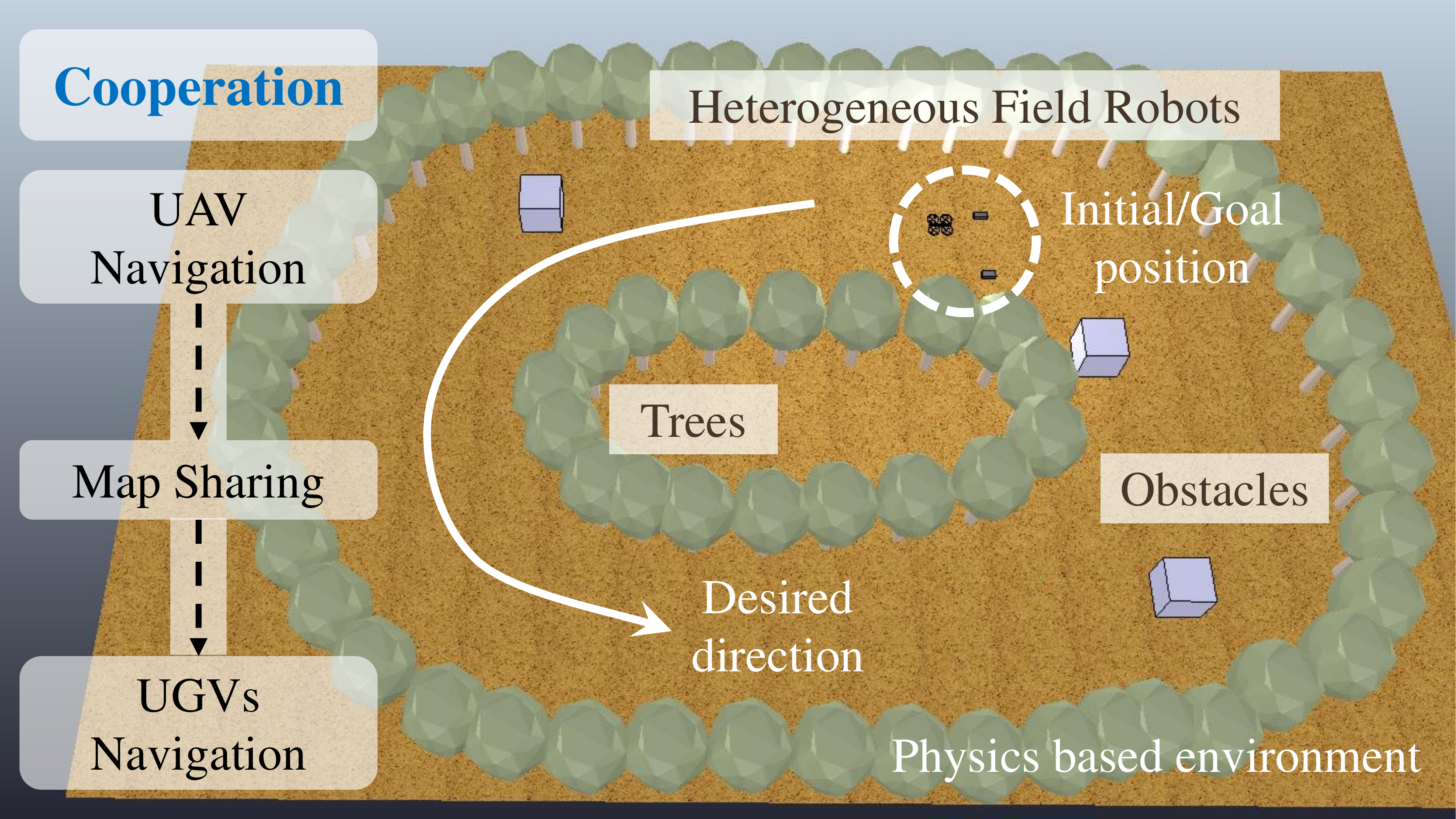}
    \caption{Simulation environments for HSHC.}
    \label{fig:sim_env}
\end{figure}
\begin{figure}[tb]
    \centering
        \includegraphics[trim={0cm 0 0cm 0},width=0.6\linewidth,page=1]{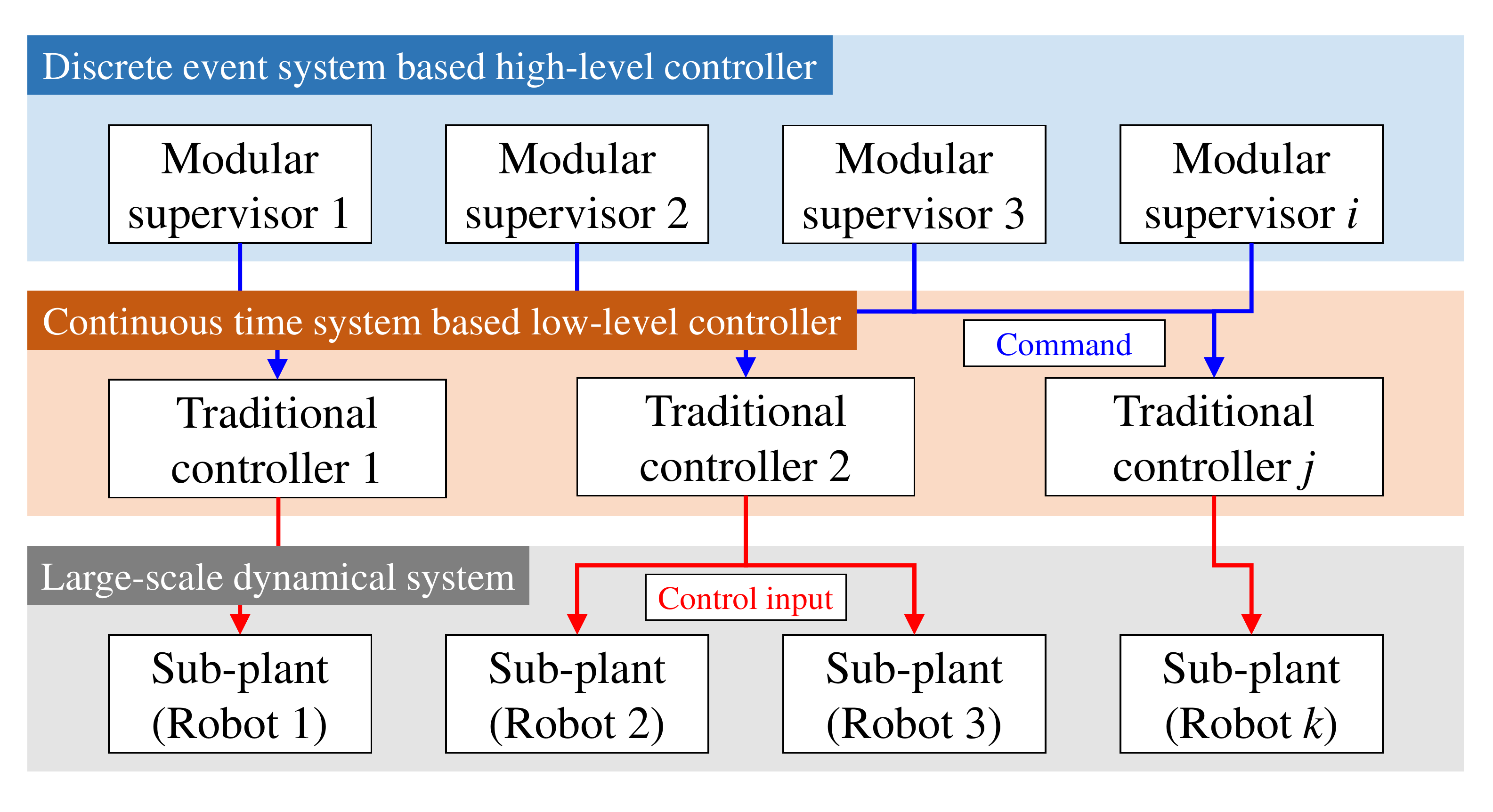}
        \caption{Hybrid systems-based hierarchical control architecture.}
\label{fig:hshc}
\end{figure}

\section{Experimental Results and Discussion}
\label{sec:5}

\subsection{Experimental Setup}
\label{sec:5.1}

\subsubsection{Environmental Setup}
\label{sec:5.1.1}
A physics-based simulator was constructed to implement and verify the designed modular supervisory controller and HSHC architecture to control heterogeneous field robots, as shown in Fig.~\ref{fig:sim_env}. The heterogeneous robot system includes one UAV and two UGVs. The virtual environment is described as an agricultural orchard. Each robot was equipped with LIDAR and camera sensors to recognize obstacles and assign a desired path to the starting point and target point through a start algorithm. The implementation of the entire system uses V-REP, a robot simulator, and MATLAB for low-level and high-level controllers. Therefore, the HSHC architecture for controlling heterogeneous field robots is shown in Fig.~\ref{fig:hshc}.
For the hybrid systems, a DES and SCT computation tool, TCT, is combined with CST-based MATLAB that sends real-time data to and receives them from the simulator. The lower controller and the modular supervisor communicate with each other through the DES-to-CTS interface and the CTS-to-DES interface including the control logic channel and the information channel and send or receive command signals through the mapping function. To evaluate the performance of the proposed HSHC system, the measured output was recorded at 50 Hz. Furthermore, to analyze the experimental results, the states of the plant, the occurrence of events, the position and velocity of the heterogeneous robot, and the VP were recorded.


\subsubsection{Experimental Task}
\label{sec:5.1.2}
In these experiments, heterogeneous robots aim for collaborative mapping and driving in an unknown field. The experimental task executed is as follows: A UAV first navigates in the agricultural environment and then shares the built map with the UGVs. Therefore, the UGVs drive based on the shared map for autonomous navigation and obstacle avoidance. 
We set up cubes and trees as obstacles, and the heterogeneous robots map and drive through onboard sensors and controllers. The following four experiments were conducted by combining the existence of obstacles and two path scenarios:
\begin{itemize}
    \item case 1: circular path with obstacles,
    \item case 2: circular path without obstacles,
    \item case 3: straight path with obstacles, and
    \item case 4: straight path without obstacles.
\end{itemize}
In this experimental task, we focused on whether heterogeneous robots perform a given cooperative task while achieving the control objectives of the modular supervisors.

\subsection{Experimental Results of Field Tasks}
\label{sec:5.2}

\begin{figure}[tb]
    \centering
    \subfigure[]
    {
        \includegraphics[trim={2.5cm 0 2.5cm 0},width=0.4\linewidth,page=1]{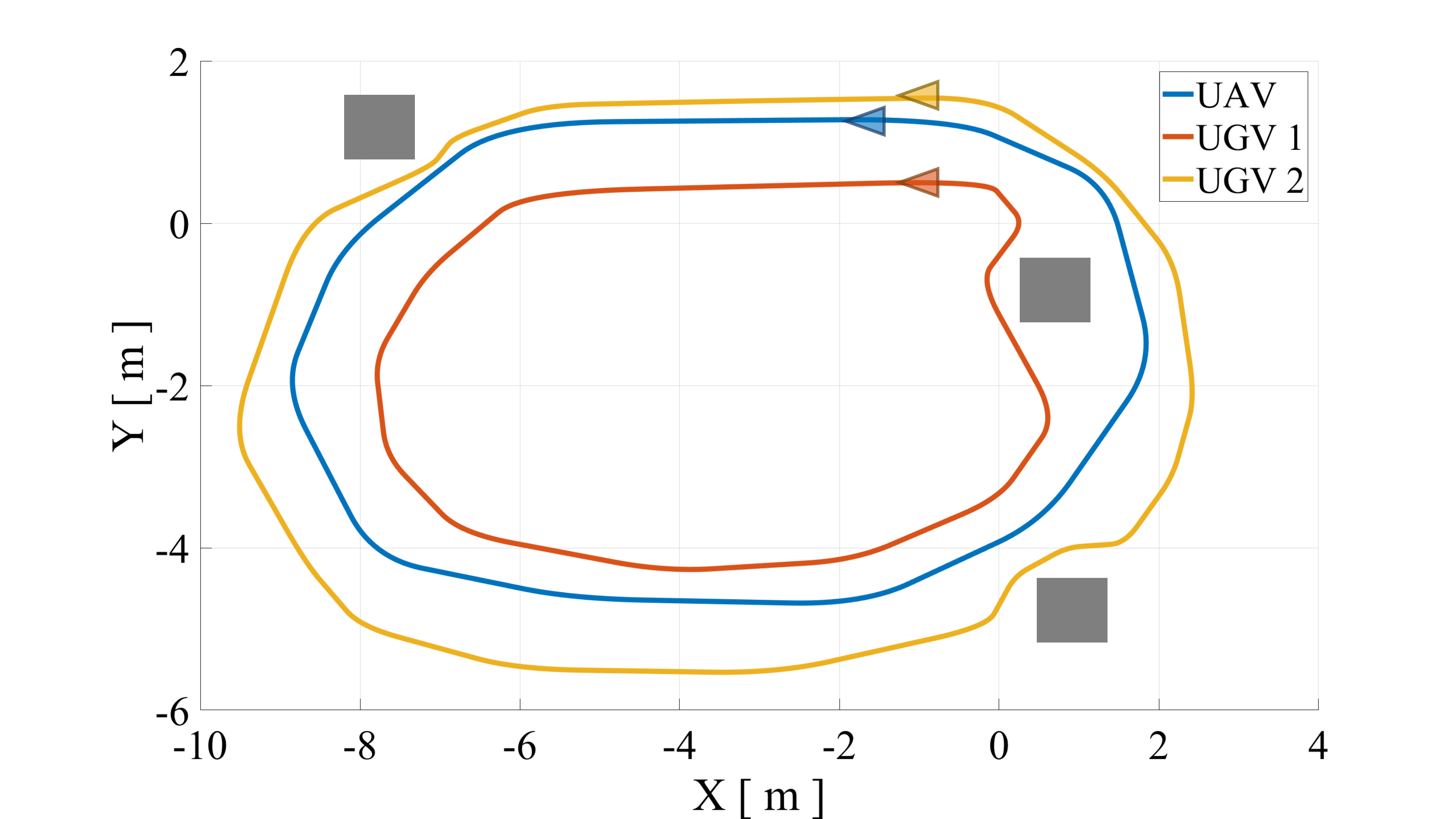}
        \label{fig:traj_1}
    }
    \\
    \subfigure[]
    {
        \includegraphics[trim={2.5cm 0 2.5cm 0},width=0.4\linewidth]{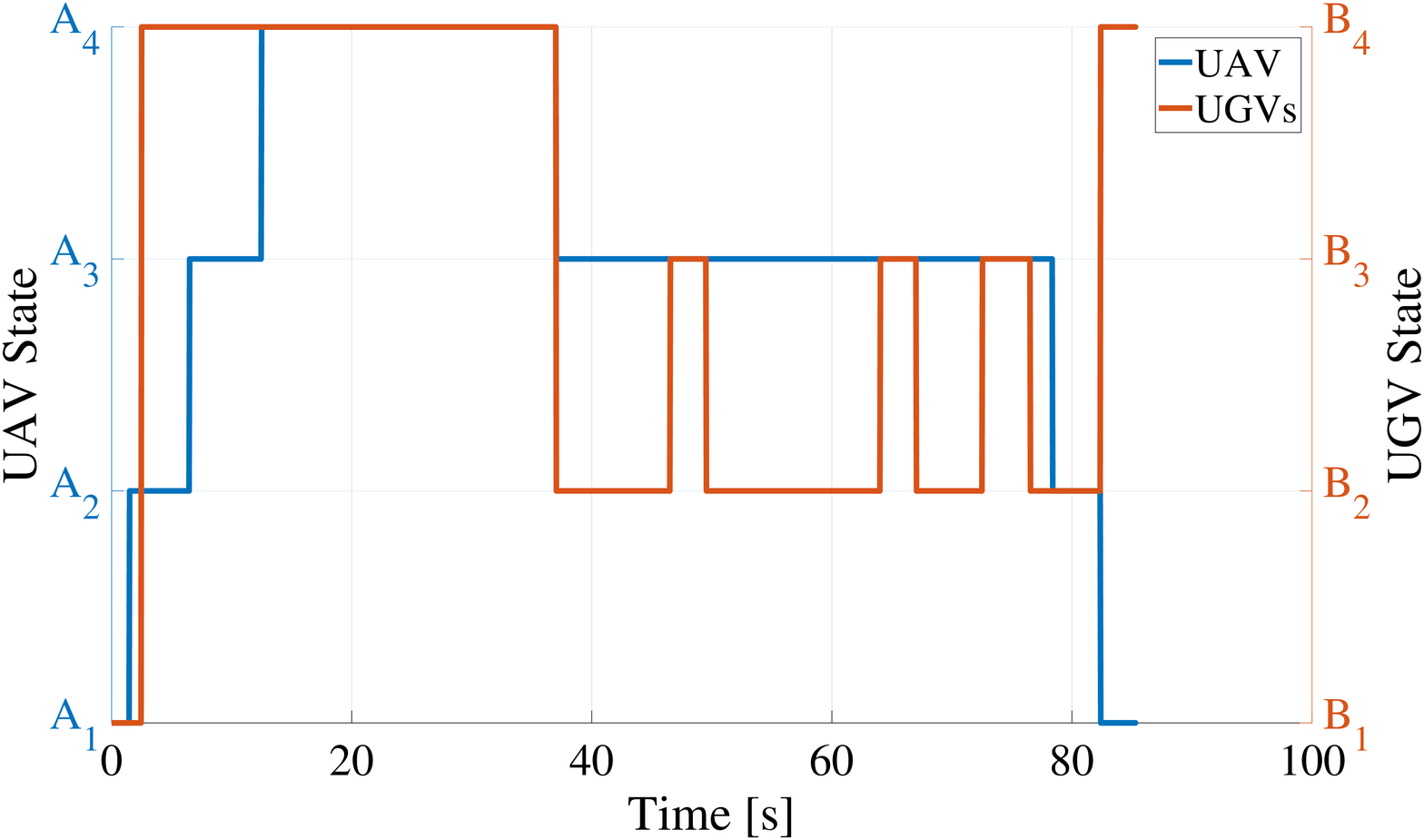}
        \label{fig:state_transition_1}
    }
     \\
    \subfigure[]
    {
        \includegraphics[trim={2.5cm 0 2.5cm 0},width=0.4\linewidth]{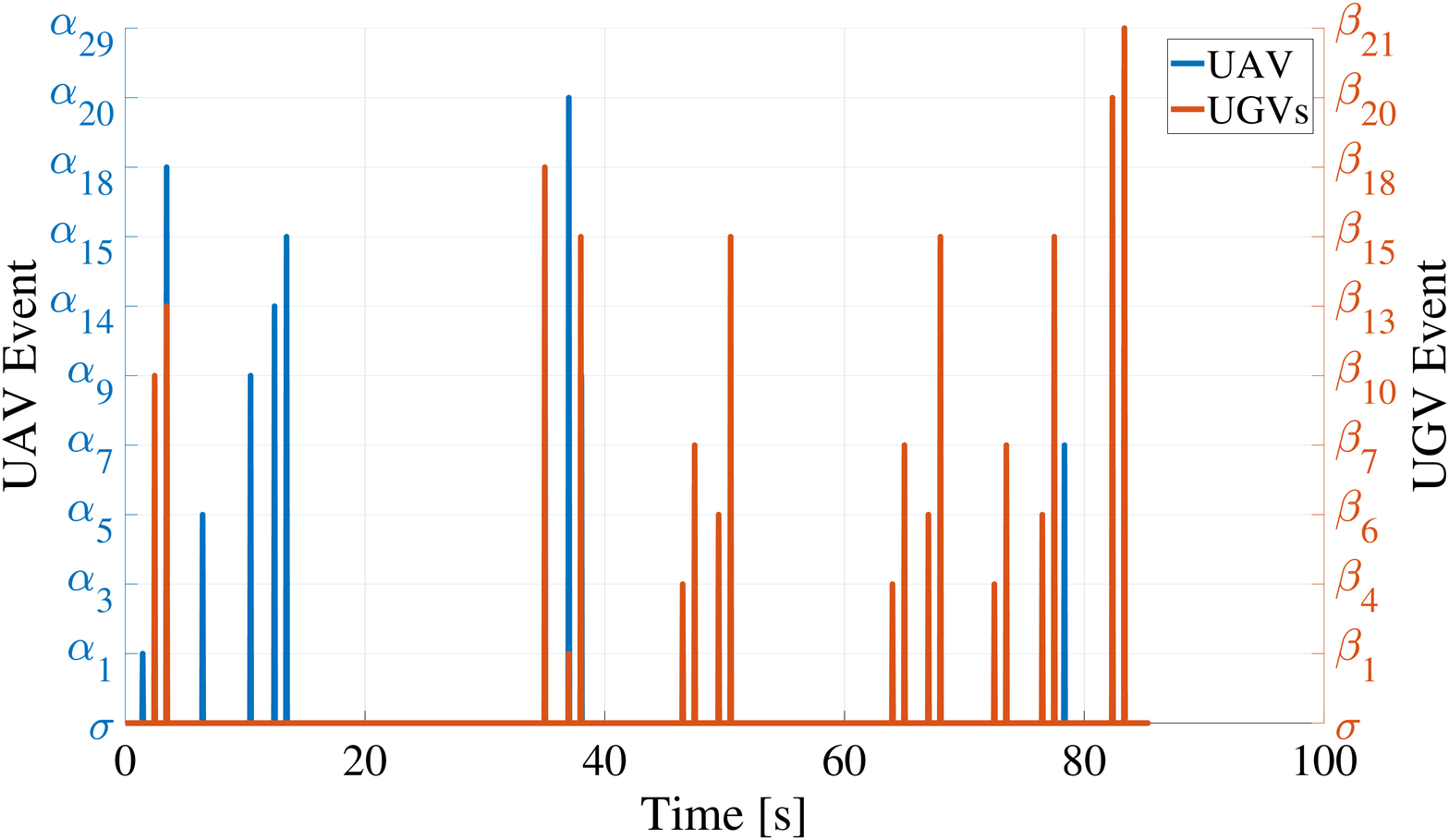}
        \label{fig:event_string_1}
    }
    \caption{Experimental results for case 1: (a) trajectory, (b) state transition, and (c) event occurrence.}
\label{fig:res_1}
\end{figure}

\begin{figure}[tb]
    \centering
    \subfigure[]
    {
        \includegraphics[trim={2.5cm 0 2.5cm 0},width=0.4\linewidth,page=1]{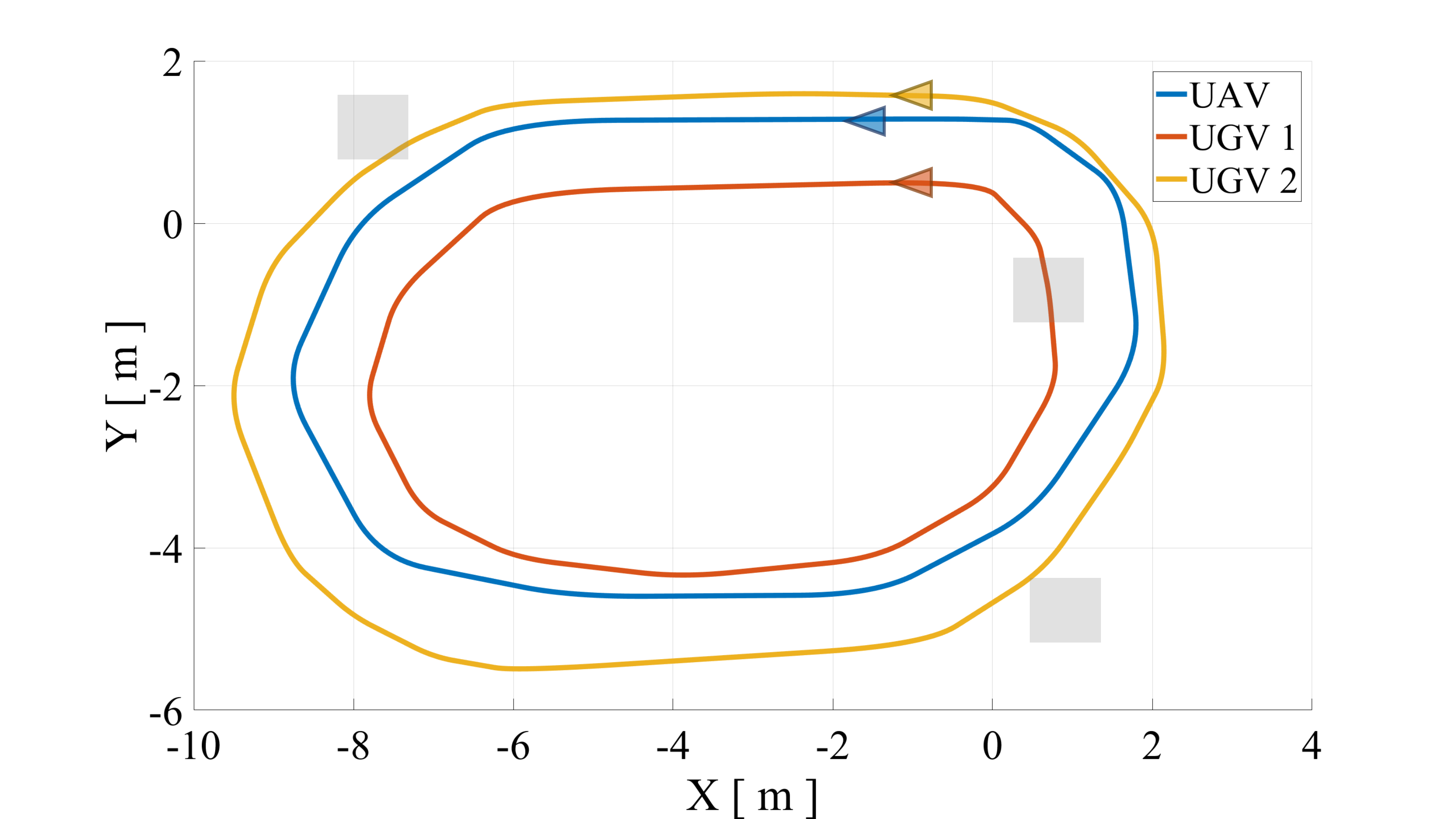}
        \label{fig:traj_2}
    }
    \\
    \subfigure[]
    {
        \includegraphics[trim={2.5cm 0 2.5cm 0},width=0.4\linewidth]{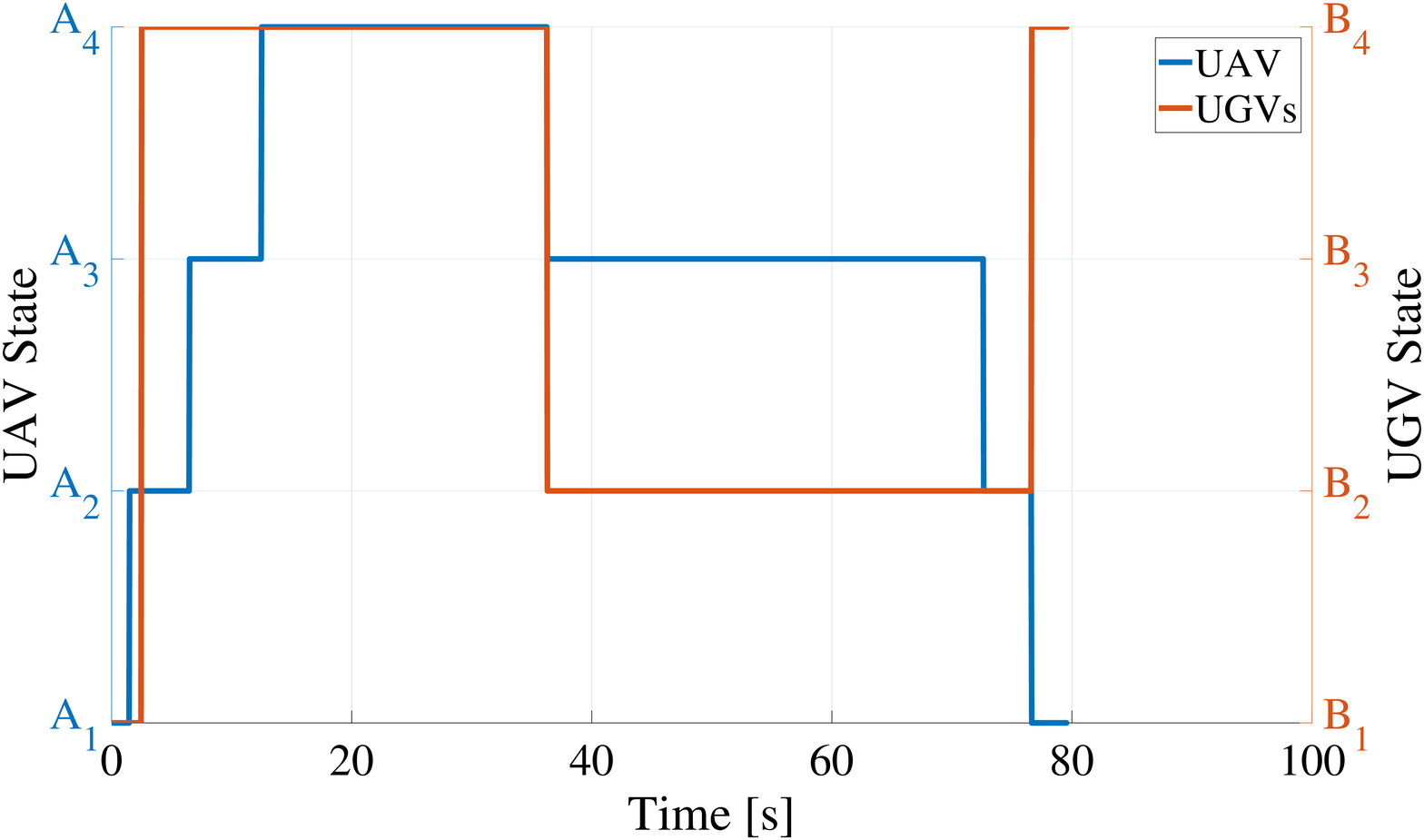}
        \label{fig:state_transition_2}
    }
     \\
    \subfigure[]
    {
        \includegraphics[trim={2.5cm 0 2.5cm 0},width=0.4\linewidth]{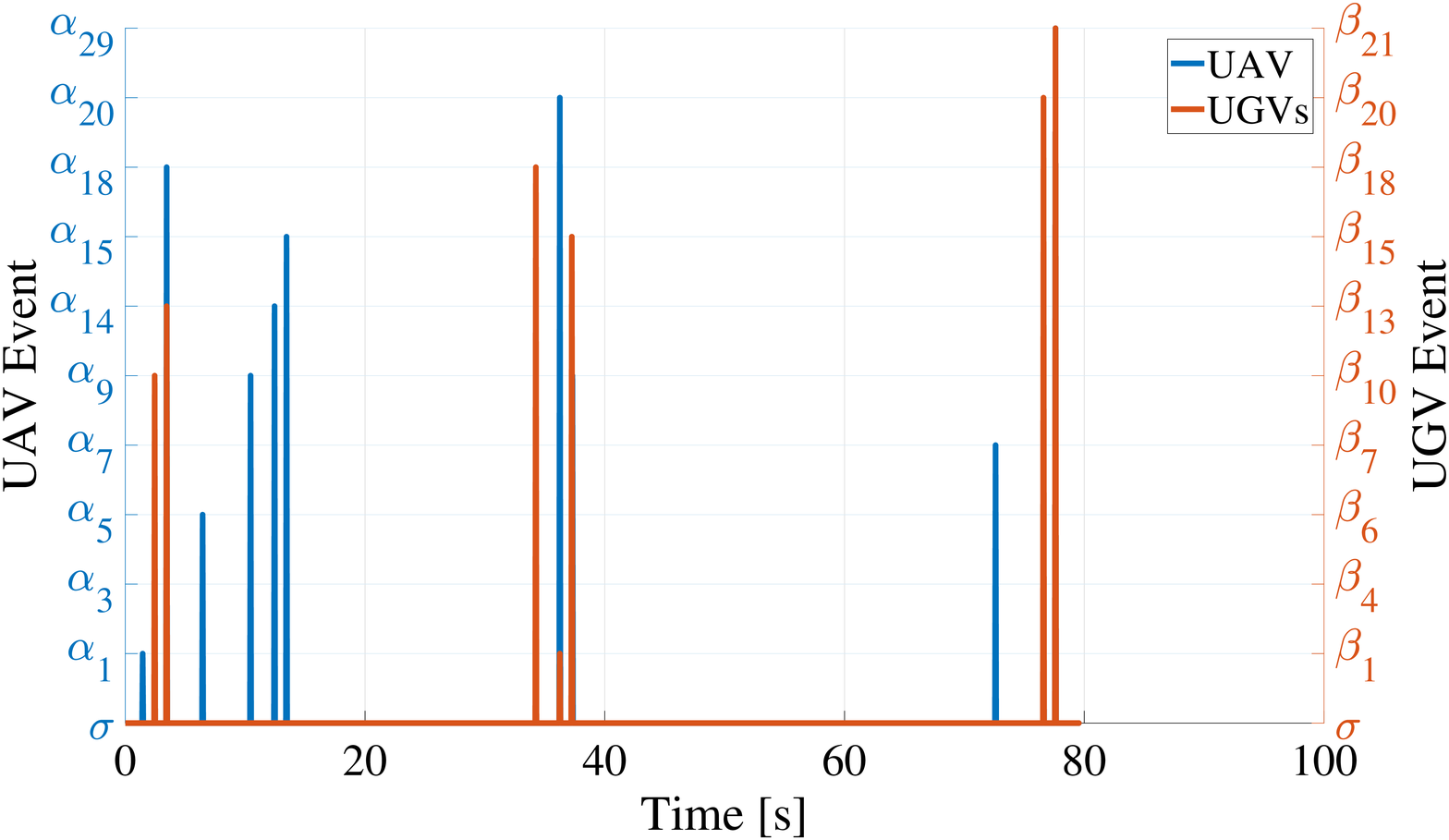}
        \label{fig:event_string_2}
    }
    \caption{Experimental results for case 2: (a) trajectory, (b) state transition, and (c) event occurrence.}
\label{fig:res_2}
\end{figure}

\subsubsection{Scenario 1 (Circular Path with and without Obstacles)}
\label{sec:5.2.1}
The experimental results of scenario 1 for cases 1 and 2 are shown in Figs.~\ref{fig:res_1} and \ref{fig:res_2}, respectively. In these figures, panels a, b, and c show the trajectories, state transitions, and observed events of the heterogeneous field robots controlled by the designed modular supervisors, respectively. Figs.~\ref{fig:traj_1} and \ref{fig:traj_2} indicate the navigated path, confirming that the UGVs avoid obstacles and do not collide with each other with the aid of the distributed swarm controller when they cooperate. Additionally, Figs.~\ref{fig:state_transition_1} and \ref{fig:state_transition_2} show the state transition of the UAV and UGVs, clearly indicating the current states of the field robots during the experiment. For example, in Fig.~\ref{fig:state_transition_1}, the UAV maintains the state of $A_4$ to perform the given mission through the arming state $A_2$ and the hovering state $A_3$ from the ideal state $A_1$. After completing the given mission, the UAV maintains the hovering state $A_3$, while the UGVs' missions are in progress, and then transitions to the arming state $A_2$ by enabling a land event when the missions of the UGVs are completed. If the mission is cleared, the UAV state changes to $A_1$ and the field experiment ends. According to the experimental result depending on the presence or absence of obstacles (i.e., comparison of cases 1 and 2 ), the red line representing the state of the UGV in Fig.~\ref{fig:state_transition_2} does not transition to $B_3$, which is a state of avoiding obstacles. However, in Fig.~\ref{fig:state_transition_1}, the UGV detects obstacles three times while driving; thus, the experimental result demonstrates that the UGV's state transitions to $B_3$ during the given mission. Furthermore, Figs.~\ref{fig:event_string_1} and \ref{fig:event_string_2} depict events that shift the behavior of the heterogeneous field robots. These systematic results are advantages of the proposed HSHC system, which ensures that it can manage large-scale dynamic systems by observing events and states to obtain the desired behavior.

\begin{figure}[tbp]
    \centering
    \subfigure[]
    {
        \includegraphics[trim={2.5cm 0 2.5cm 0},width=0.4\linewidth,page=1]{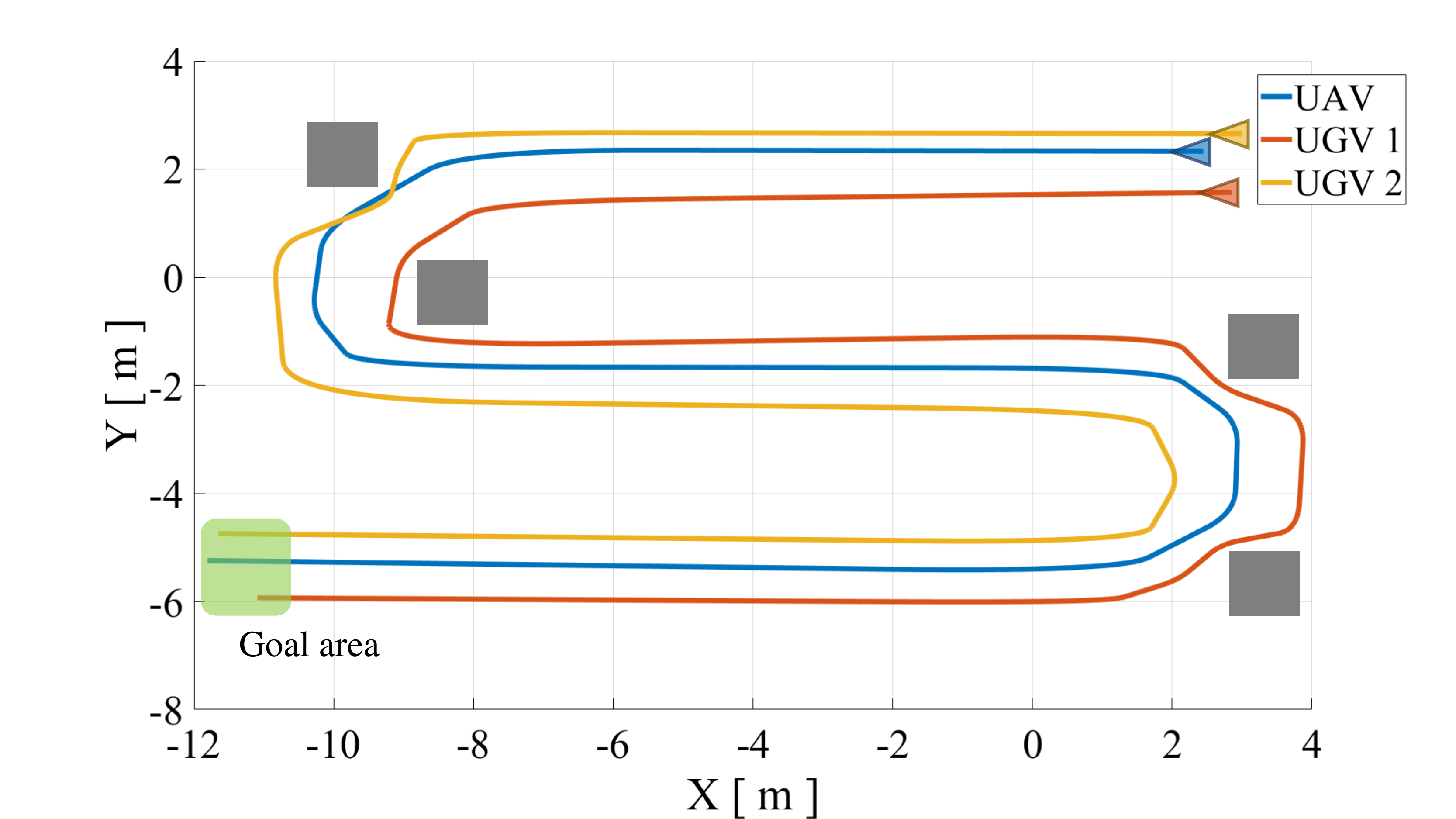}
        \label{fig:traj_3}
    }
    \\
    \subfigure[]
    {
        \includegraphics[trim={2.5cm 0 2.5cm 0},width=0.4\linewidth]{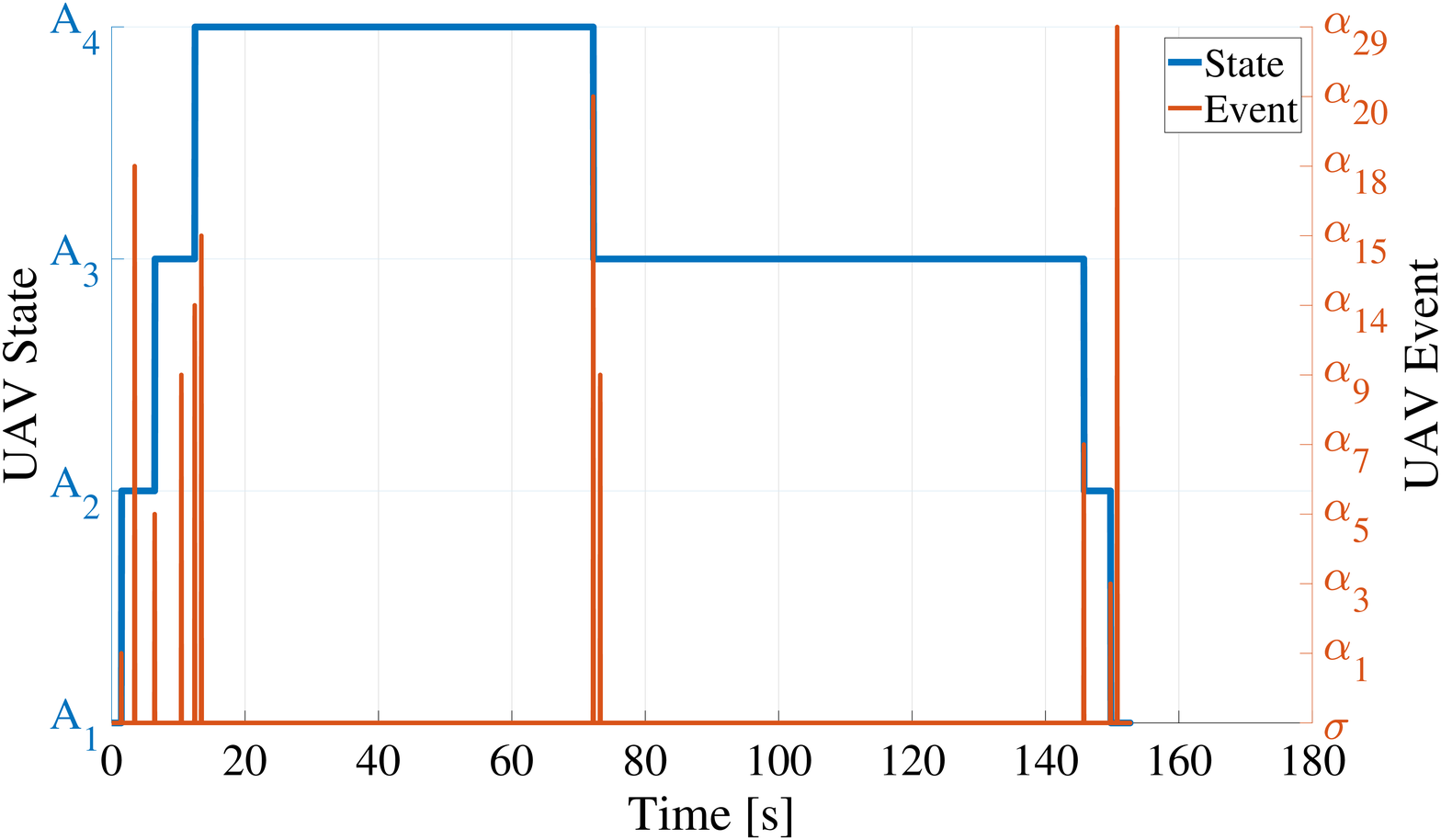}
        \label{fig:state_transition_3}
    }
     \\
    \subfigure[]
    {
        \includegraphics[trim={2.5cm 0 2.5cm 0},width=0.4\linewidth]{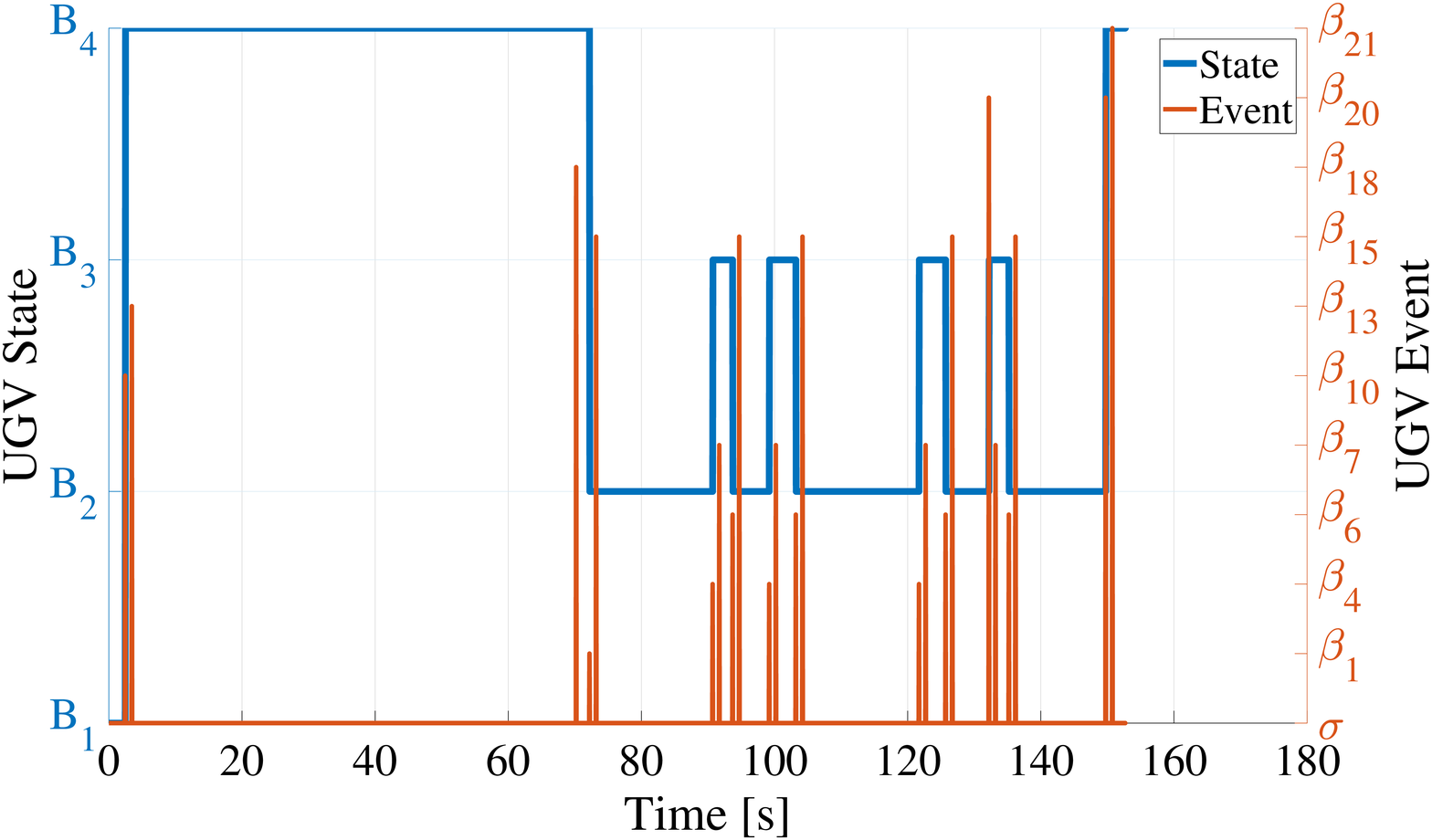}
        \label{fig:event_string_3}
    }
    \caption{Experimental results for case 3: (a) trajectory, (b) UAV, and (c) UGV.}
\label{fig:res_3}
\end{figure}

\begin{figure}[tbp]
    \centering
    \subfigure[]
    {
        \includegraphics[trim={2.5cm 0 2.5cm 0},width=0.4\linewidth,page=1]{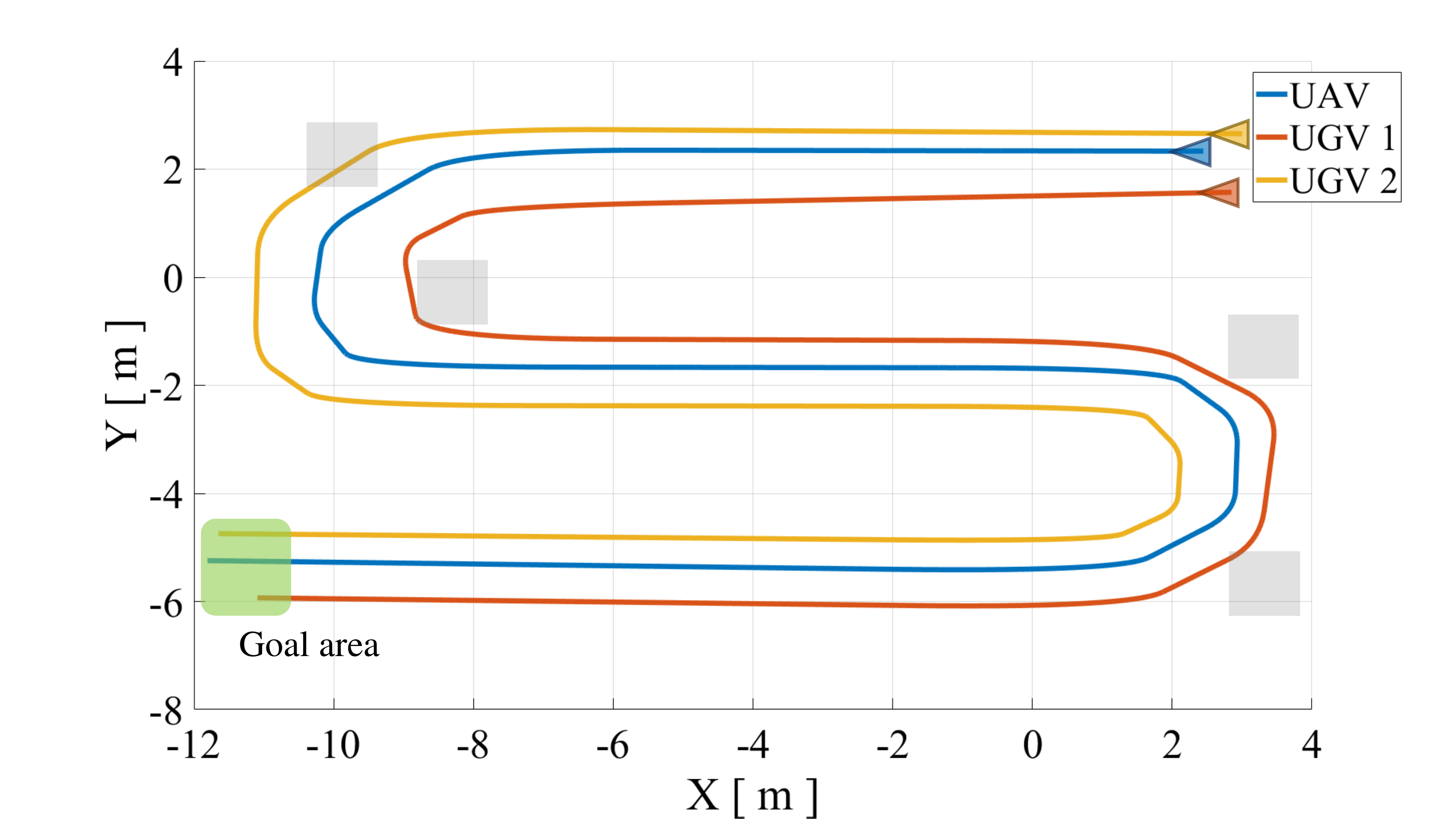}
        \label{fig:traj_4}
    }
    \\
    \subfigure[]
    {
        \includegraphics[trim={2.5cm 0 2.5cm 0},width=0.4\linewidth]{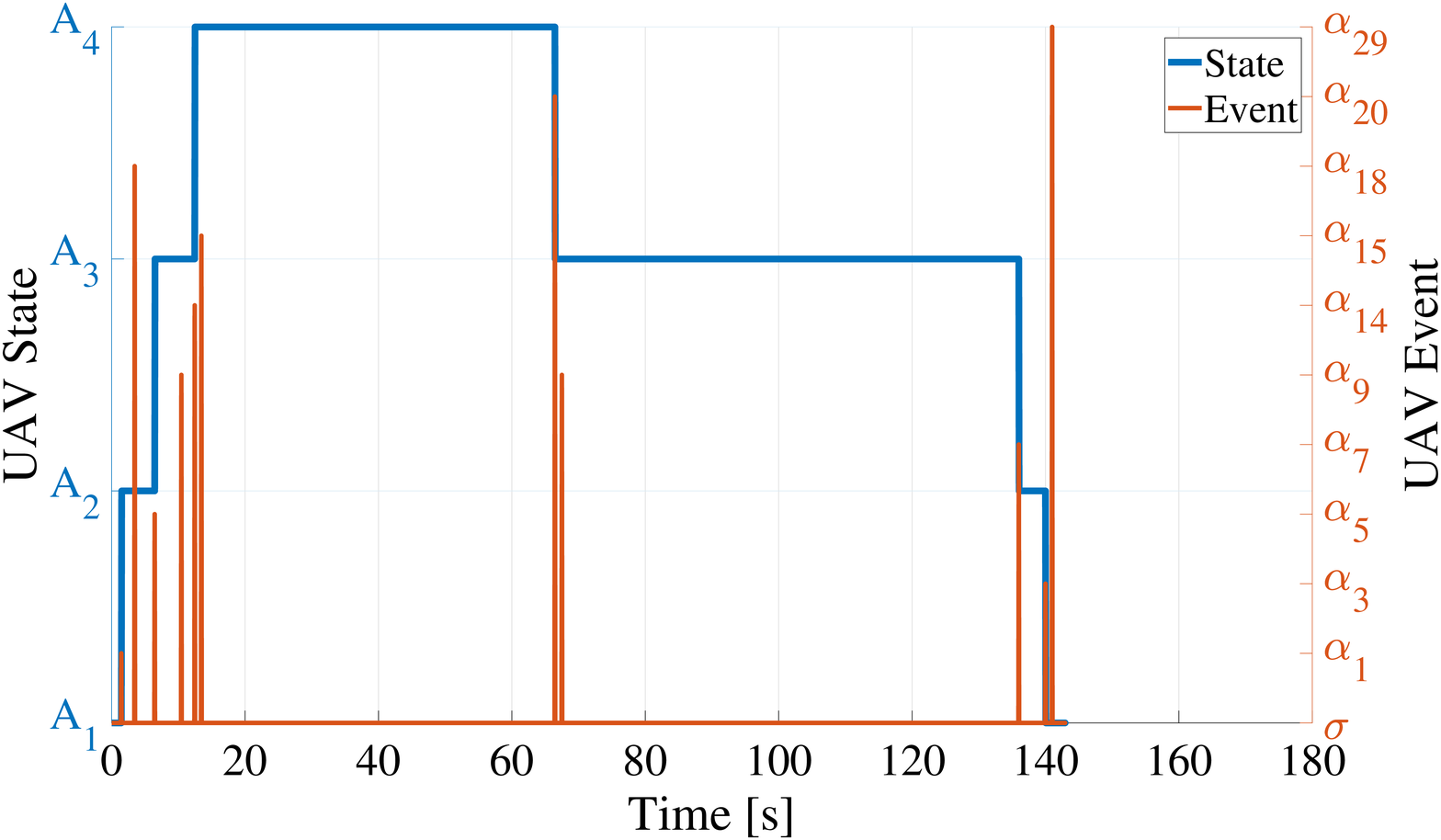}
        \label{fig:state_transition_4}
    }
     \\
    \subfigure[]
    {
        \includegraphics[trim={2.5cm 0 2.5cm 0},width=0.4\linewidth]{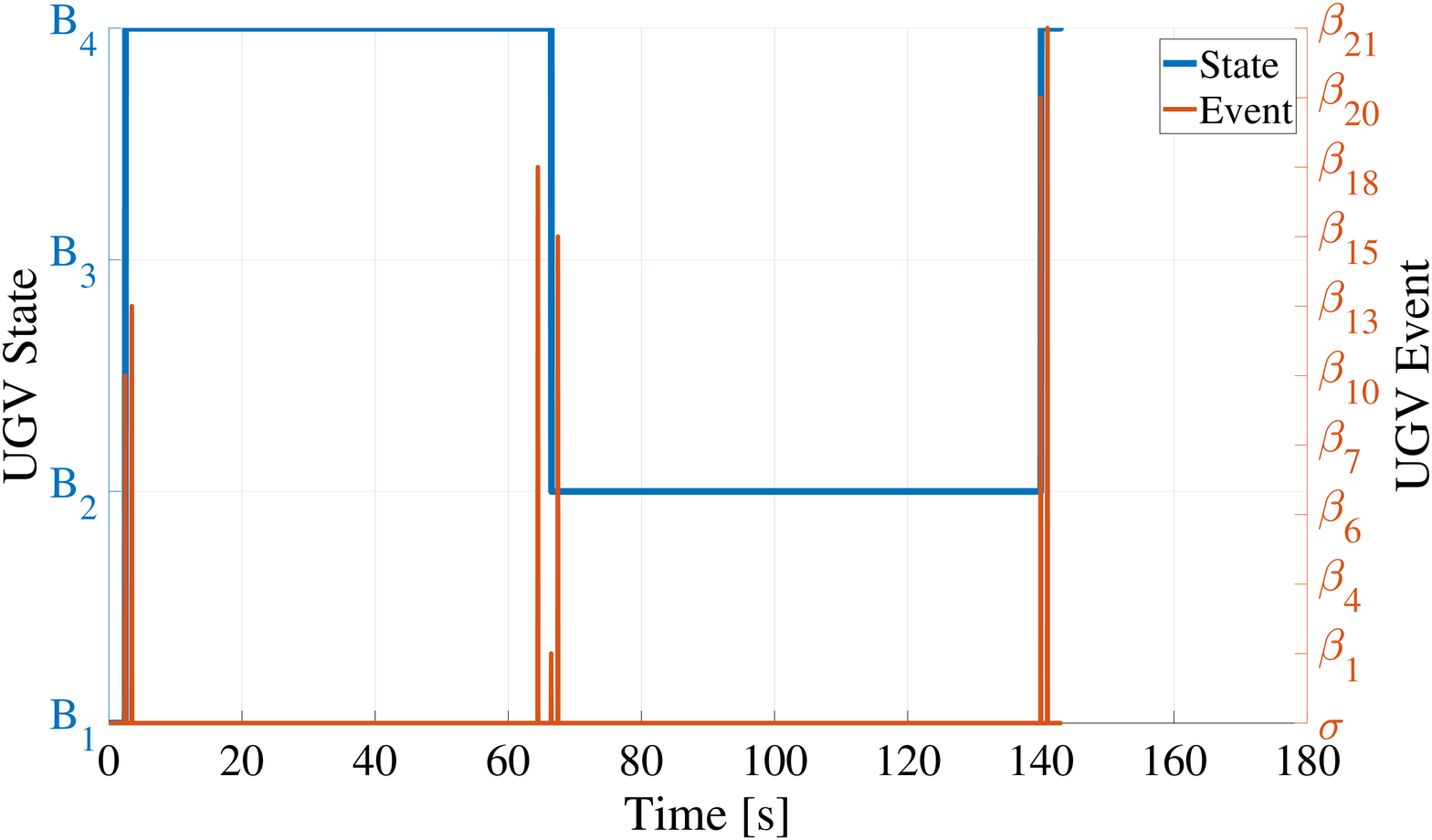}
        \label{fig:event_string_4}
    }
    \caption{Experimental results for case 4: (a) trajectory, (b) UAV, and (c) UGV.}
\label{fig:res_4}
\end{figure}

\subsubsection{Scenario 2 (Straight Path with and without Obstacles)}
\label{sec:5.2.2}
Figs.~\ref{fig:res_3} and \ref{fig:res_4} depict the experimental results for the heterogeneous field robot systems for scenario 2. Figs.~\ref{fig:traj_3} and \ref{fig:traj_4} show that the field robots follow a given path, avoid obstacles, and form the desired formation. These figures are the result of the system behaviors under the obstacle avoidance controller, formation controller, path-following controller introduced in Section~\ref{sec:2}, and the modular supervisors designed in Section~\ref{sec:3}. Figs.~\ref{fig:state_transition_3} and \ref{fig:state_transition_3} show the state transitions and event occurrences of the UAV plant for each case. The state transitions of the UAV between cases 3 and 4 are not very different because the UAV flies over obstacles. For the control purposes of this experiment, we verified whether the specifications designed in Section~\ref{sec:4.3} were met, and found that the UAV, the leader robot, performs its mission according to the desired process. For example, Figs. \ref{fig:state_transition_3} and \ref{fig:state_transition_3} show that the UAV takes off and maintains a hovering state before it performs its mission. Additionally, the UAV satisfies the specification that the mission should be assigned while arming. Figs.~\ref{fig:event_string_3} and \ref{fig:event_string_4} show the state transitions and event occurrences of the UGVs' plant for cases 3 and 4 in scenario 2, respectively. In the case of UGVs that navigate the unknown field environment based on the UAV-generated map, these results ensure that the states of obstacle avoidance and navigation are originated from the events that detect obstacles and free space. In detail, these graphs demonstrate that the state of the UGV is changed from B2 to B3 when an event that detects an obstacle occurs. However, in Fig.~\ref{fig:event_string_4}, the UGVs continue to cooperatively work because there are no obstacle environments. In other words, the modular supervisors manage the controllable events to ensure that the plants have the desired behavior while observing eligible events.

\label{sec:5.2.3}
\begin{figure}[tb]
    \centering
    \includegraphics[width=0.6\linewidth,page=1]{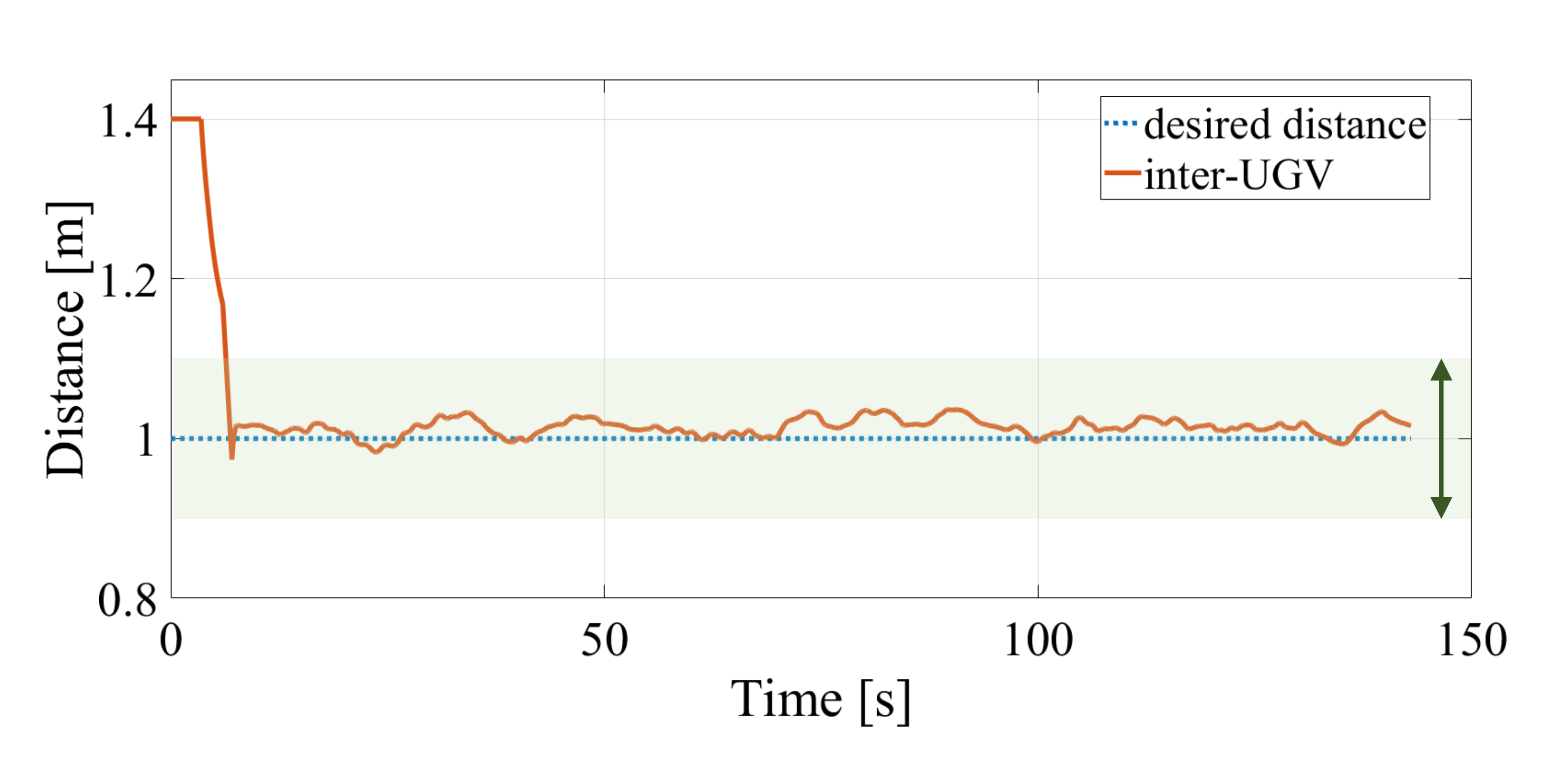}
    \caption{Relative distance between UGVs for case 4.}
    \label{fig:rel_dist}
\end{figure}

\subsubsection{Formation Control of Heterogeneous Field Robots}

Fig.~\ref{fig:rel_dist} shows the relative distance between UGVs during the experiments, and this result is affected by the formation controller introduced in Section~\ref{sec:2}. Theoretically, the desired relative distance is 1 m; however, it is difficult to maintain this distance because of the measurement error and disturbances. Therefore, the formation of UGVs is controlled by a potential function with a specific range that is designed to maintain a distance of 0.9--1.1 m in this experiment. From Figs.~\ref{fig:event_string_4} and~\ref{fig:rel_dist}, we can observe that the UGVs form the desired formation after the network is connected. Therefore, heterogeneous field robot systems composed of multiple follower robots can collaborate through the proposed HSHC architecture. In other words, when an event indicating a network connection occurs, the controllable event, the formation control input, is enabled by the modular supervisor, and then the low-level controller regulates the heterogeneous robot.

\subsection{Discussion}
\label{sec:5.3}

Multiple robots based systems for autonomous field tasks must be guaranteed for scalability and responsiveness. For example, when a humanoid robot is added to the existing robot team consisting of UAV and UGV to collaborate, modeling and controller design of the robot system are inevitable and challenging. However, the proposed HSHC approach has the advantages of modularity, scalability, reactivity, and explainability in comparison with classical control methods. In other words, modeling a subplant $\mathcal{G}_{sub,i}$ for a new environment and designing the specifications $H_i$ to obtain a modular supervisor $\mathcal{S}_j$ creates a novel control system while systematically observing the states and events. Remodeling and redesign of large-dynamic systems are identical processes, but the HSHC architecture is a more prompt, modular, and systematic approach. As another example, to obtain the dynamic model of the field robots equipped with a specific sensor (e.g., laser, sonar, and Lidar), the robot models $\mathcal{G}_A$ and $\mathcal{G}_B$ are synthesized with a sensor model $\mathcal{G}_{sensor}$ by considering the states $\mathcal{E}$ and events $\Omega$ of the sensor~\cite{gonzalez2017supervisory}. Therefore, considering this scalability, HSHC-based heterogeneous field robot teams are guaranteed applicability and adaptability for dynamic environments.

However, the SCT-based HSHC system also faces various drawbacks and challenges. This approach is intuitive and plain because it models systems and designs controllers based on formal languages. Besides, extremely complex systems have extensive states, events, and state transition functions. Therefore, graphical tools must be introduced to represent the formal languages clearly, and it is complicate to debug and apply HSHC for the implementation without an event generator. Moreover, the low-level controllers for dynamic systems must be established to apply HSHC architecture. Nevertheless, if the above-mentioned components for control and implementation are constructed, the HSHC approach is efficient to manage the heterogeneous multi-robot systems. In the challenges, there are centralized, decentralized, distributed, and hierarchical approaches to the design of supervisory controllers~\cite{wonham2015supervisory}. Additionally, depending on the type of control method, the optimal, robust, adaptive supervisory controller can be considered for the nonlinear hybrid dynamical systems~\cite{buss2002nonlinear}. The integration of artificial intelligence with HSHC and SCT for learning also poses a future challenge~\cite{kang2019machine}. More detailed and advanced research that considers the observability, failure diagnosis, and parallel executions of hybrid systems such as behavior trees~\cite{marzinotto2014towards}, is of utmost importance, and we intend to pursue these challenges in our future work.

\section{Conclusions}
\label{sec:6}
We proposed an HSHC architecture for the cooperation of heterogeneous field robot team used hybrid automata and the formal modeling approach for the hybrid systems consisted of CTS and DES. Their performance and effectiveness were verified and evaluated through the heterogeneous field robot system consisted of one UAV and two UGVs. For the modeled large-scale dynamic plant, we designed a legal behavior (specifications) to achieve the control objectives. Based on these specifications, modular supervisory controllers were synthesized. We also evaluated the nonblocking, nonconflictness, and controllability to verify that the modular supervisors were proper. All modular supervisors can control and manage the entire plant exactly same with a centralized supervisor. To evaluate our proposed HSHC approach and develop a supervisory control system, we conducted its implementation and performed experiments on it in a physics-based simulator. The experimental results confirmed that the heterogeneous field robots achieve the given control objectives, and systematic results, such as the system behavior and the event string are presented.

Based on the results obtained in field experiments and the previous results from our studies~\cite{8834867} (please note that a field experiment was already done based on some parts of HSHC in~\cite{8834867}), the feasibility and effectiveness of our proposed HSHC system were sufficiently validated. Therefore, hybrid systems and SCT are more efficient and scalable than traditional control systems to control large-scale dynamic systems, such as heterogeneous robot systems. For example, when an additional field robot is included in the entire plant, the subplant is modeled with hybrid automata and synthesized in the plant, and the specifications for the added model are designed to obtain a modular supervisor, which is effective for addressing complexity, modularity, and reactivity. In our future work, we intend to apply more elaborate scenarios and implement the HSHC architecture fully and the heterogeneous field robots in real outdoor environments to evaluate their practicality. Implementing this approach can be of significant value to advanced and autonomous field robot systems in the future (e.g., large-scale environmental sampling and monitoring and smart farming), and research on these HSHC systems are expected to proceed in various directions, such as multitasking, multiareas, observability, and failure diagnosis.

\bibliographystyle{unsrtnat}
\bibliography{references}  

\end{document}